\documentclass[10pt,journal,compsoc]{IEEEtran}

\usepackage{ifpdf}
\ifCLASSOPTIONcompsoc
  \usepackage[nocompress]{cite}
\else
  \usepackage{cite}

\fi

\ifCLASSINFOpdf
\else
\fi

\usepackage{amsmath}
\usepackage{amssymb}
\usepackage{bm}
\usepackage{array}
\usepackage{algorithm}
\usepackage{comment}
\usepackage{algpseudocode}
\usepackage{graphicx}
\usepackage{xcolor}
\usepackage{booktabs}
\usepackage{threeparttable}
\usepackage{multirow}
\usepackage{arydshln}
\usepackage{url}

\newcommand{\etal}{\textit{et al}.}
\newcommand{\ie}{\textit{i}.\textit{e}.}

\newcommand\etc{etc\@ifnextchar.{}{.\@}}

\newcommand{\Rnum}[1]{\uppercase\expandafter{\romannumeral #1\relax}}

\begin{document}

\title{Disentangled Human Body Embedding Based on Deep Hierarchical Neural Network}

\author{Boyi~Jiang,~
        Juyong~Zhang$^\dagger$,
	 Jianfei~Cai,
        and~Jianmin~Zheng
\IEEEcompsocitemizethanks{\IEEEcompsocthanksitem B. Jiang and J. Zhang are with School of Mathematical Sciences, University of Science and Technology of China. 
\IEEEcompsocthanksitem J. Cai is with Faculty of IT, Monash University.
\IEEEcompsocthanksitem J. Zheng is with School of Computer Science and Engineering, Nanyang Technological University.}
\thanks{$^\dagger$Corresponding author. Email: {\texttt{juyong@ustc.edu.cn}}.}
}

\markboth{IEEE Transactions on Visualization and Computer Graphics}
{Jiang \MakeLowercase{\textit{et al.}}: Learning 3D Human Body Embedding}

\IEEEtitleabstractindextext{%
\begin{abstract}
Human bodies exhibit various shapes for different identities or poses, but the body shape has certain similarities in structure and thus can be embedded in a low-dimensional space. This paper presents an autoencoder-like network architecture to learn disentangled shape and pose embedding specifically for the 3D human body. This is inspired by recent progress of deformation-based latent representation learning.
To improve the reconstruction accuracy, we propose a hierarchical reconstruction pipeline for the disentangling process and construct a large dataset of human body models with consistent connectivity for the learning of the neural network.
Our learned embedding can not only achieve superior reconstruction accuracy but also provide great flexibility in 3D human body generation via interpolation, bilinear interpolation, and latent space sampling.
The results from extensive experiments demonstrate the powerfulness of our learned 3D human body embedding in various applications.
\end{abstract}

}

\maketitle
\IEEEdisplaynontitleabstractindextext
\IEEEpeerreviewmaketitle

\section{Introduction}

This paper considers the problem of learning a parametric 3D human body model, which can map a low-dimensional latent representation into a high-quality 3D human body mesh via a deep hierarchical neural network. Parametric human body models have a wide range of applications in computer graphics and computer vision. Examples of such applications are body tracking~\cite{xu2018monoperfcap,yu2018doublefusion}, body reconstruction~\cite{alldieck2018video,bogo2016keep,loper2014mosh} and pose estimation~\cite{omran2018neural,kanazawa2018end}. However, building an expressive and reliable parametric body model is challenging. This is because the human body has abundant variations due to many factors, such as gender, ethnicity, and stature. In particular, different poses may introduce significant deformations of the body, which are difficult to model by conventional linear techniques such as principal component analysis (PCA).

The state-of-the-art work SMPL (skinned multi-person linear model)~\cite{loper2015smpl} separates human body variations into shape-related variations and 3D pose variations. The shape-related variations are modeled by a low dimensional linear shape space with shape parameters. The 3D pose variations are handled by a skeleton skinning method with pose parameters derived from 3D joint angles. SMPL has a clear pose definition and can express different human poses of large scales. The parameter-to-mesh computation in SMPL is fast and robust. However, the reconstruction accuracy of the skeleton skinning method relies on the linear shape space of neutral body shapes. The skinning weights of SMPL are shared for different neutral shapes of different identities, which further restricts its reconstruction ability. To capture the pose-dependent deformations and reduce skinning artifacts around joints, SMPL introduces independent pose-related blend shapes as the complement for the intrinsic shape space defined by shape parameters. The pose parameters of SMPL explicitly define the movements of the human skeleton and are very suitable for character animation editing. Due to the excess expression ability of the pose parameters, specific human body pose prior is always needed for applications to avoid occurrences of unnatural body meshes. \cite{bogo2016keep,alldieck2017optical} used some pose prior constraints like joint angle assignment range and self-intersection penalty energy to generate plausible body shapes. \cite{kanazawa2018end} adopted a network discriminator to judge whether the generated pose parameters obey the distribution of human motion during training. Different from these works, we introduce a novel disentangled body representation that can achieve better accuracy in body shape reconstruction and whose pose latent parameters encode the prior of human pose distribution to some extent.

With the advance of deep learning, the encoder-decoder based architecture has demonstrated its capability of extracting latent representations of face geometry~\cite{bagautdinov2018modeling,ranjan2018generating,jiang2019disentangled}. Compared with face shape, human body shape is more complicated as it contains many joints and very complex movements. Therefore, directly extending the neural network-based method for face shape to human body shape cannot achieve good performance. Recently, Litany~\etal~\cite{litany2017deformable} proposed a graph convolution-based variational autoencoder for 3D body shape, which directly uses the Euclidean coordinates as the vertex feature and encodes the whole shape without disentangling identity and posture attributes. However, Euclidean-domain based encoder-decoder architecture may produce non-natural deformation bodies from latent embedding. Different from Euclidean coordinates, a mesh deformation representation called ACAP (as consistent as possible) introduced in~\cite{gao2017sparse} can handle arbitrarily large rotations in a stable way and has great interpolation and extrapolation properties. The recent studies in~\cite{tan2017mesh,tan2017variational} show that learning on deformation features with autoencoder or VAE~\cite{kingma2013auto} can achieve more powerful latent representation. However, \cite{tan2017mesh,tan2017variational} are designed for learning latent representation of general 3D shapes. When applied to 3D human body modeling, they only provide one latent embedding that entangles both shape and pose variations, which is not sufficient for practical uses.

Therefore, in this paper, we propose to utilize the neural network to learn two disentangled latent representations from ACAP features: one for shape variations and the other for pose variations, both of which are specifically designed and learned for the human body modeling. Moreover, a coarse-to-fine reconstruction pipeline is integrated into the disentangling process to improve the reconstruction accuracy. Our major contributions are twofold:
\begin{itemize}
\item We propose a general framework based on variational autoencoder architecture for learning disentangled shape and pose latent embedding of 3D human body. Our framework introduces a hierarchical representation design. The basic transformation module has great design freedom.
	
\item Learning on ACAP features~\cite{gao2017sparse} requires mesh data to have the same connectivity while existing human body datasets do not satisfy this requirement. To address this issue, we re-mesh a large set of meshes from multiple existing human body datasets into standard connectivity via a novel non-rigid registration method and construct a new large scale human body dataset. The dataset consists of over 5000 human body mesh models with the same connectivity, where each identity has a standard or neutral pose.\footnote{Our full framework is available at \url{https://github.com/Juyong/DHNN_BodyRepresentation}.}
\end{itemize}
	
We have conducted extensive experiments, including various applications. The experimental results demonstrate the powerfulness of our learned 3D human body embedding in terms of modeling accuracy, generation flexibility, etc.

\section{Related Work}
\textbf{Human shape models.} Human body shape is often constructed and represented via its shape variations~\cite{allen2003space,seo2003automatic,yang2014semantic,pishchulin2017building}. For example, Anguelov \etal~\cite{anguelov2005scape} proposed to process shape completion by computing the deformation of triangles between the template and target meshes. Performing PCA on the transformation matrices further yields more robust results.
Allen \etal~\cite{allen2003space} and Seo \etal~\cite{seo2003automatic} applied PCA to mesh vertex displacements to characterize the non-rigid deformation of human body shapes. Moreover, Allen \etal~\cite{allen2003space}  constructed a correspondence between a set of semantic parameters of body shapes and PCA parameters by linear regression, which facilitates the manipulation of human body shapes. Zhou \etal~\cite{zhou2010parametric} used a similar idea to semantically reshape human bodies from a single image. To extract more local and fine-grained semantic parameters from body shape representation, Yang \etal~\cite{yang2014semantic} introduced local mapping between semantic parameters and per triangle deformation matrix, which provides precise semantic control of human body shapes.

\textbf{Human pose models.} To represent human shape with poses, skeleton skinning is often used, which can directly compute positions of vertices on the body shape. Allen \etal~\cite{allen2006learning} proposed to learn skinning weights for corrective enveloping and solve a highly nonlinear equation to find the relation among pose, skeleton, and skinning weights. Joo~\etal~\cite{joo2018total} stitched hand, face, and body models together to obtain an expressive model that can capture the motion of humans. SMPL~\cite{loper2015smpl} explicitly defines body joints, uses the skeleton to represent body pose, and computes vertex positions with the standard skinning method. Hesse~\etal~\cite{hesse2019learning} followed SMPL's design and learned a statistical 3D infant body model from sequences of incomplete, low-quality RGB-D images of freely moving infants. Pavlakos~\etal~\cite{pavlakos2019expressive} expanded SMPL to capture the hand pose and facial expression with a unified representation.

\textbf{Deformation-based models.} Mesh deformations have been used to analyze 3D human body shape and pose~\cite{anguelov2005scape,chen2013tensor,freifeld2012lie,hasler2009statistical,hirshberg2012coregistration,hasler2010multilinear}. The most representative work is SCAPE~\cite{anguelov2005scape}, which analyzes body shape and pose deformation in terms of the deformation of triangles with respect to a reference mesh. The deformation representation can encode detailed shape variations, but an optimization process is required to obtain the mesh from the deformation representation. The conversion usually causes some time, which constrains it from real-time applications~\cite{weiss2011home}. Chen \etal ~\cite{chen2016realtime} extended the SCAPE~\cite{anguelov2005scape} approach for real-time reconstruction of an animating human body. Jain \etal~\cite{jain2010moviereshape} used a common skinning approach for modeling pose-dependent surface variations instead of using per-triangle transformation, which makes the pose estimation much faster than SCAPE~\cite{anguelov2005scape}. 

\textbf{Deep learning for geometric representation.}
Bagautdinov~\etal\cite{bagautdinov2018modeling} introduced a ladder VAE architecture to effectively encode face shape in different scales, which can achieve high reconstruction accuracy. Anurag~\etal\cite{ranjan2018generating} defined upsampling and downsampling operations on the face mesh and used graph structure convolution to encode latent representation, which can obtain high reconstruction accuracy even for extreme facial expressions. The method proposed in \cite{jiang2019disentangled} disentangles identity and expression attributes with two VAE branches and then fuses them back to the input mesh. By exploiting the strong non-linear expression capability of neural network and a deformation representation, the method outperforms previous methods in
the decomposition of facial identity and expression. Hamu~\etal~\cite{ben2019multi} introduced a 3D shape generative model for genus-zero shapes and adopted a novel 3D shape tensor representation to make it suitable for arbitrary connectivity. However, the lack of disentangled representation restricts its wide application. Higgins \etal~\cite{higgins2017beta} proposed a novel strategy to automatically learn disentangled latent representations from raw data in a completely unsupervised manner.

\textbf{Deformation representation.}
Geometric representation based on Euclidean coordinates is not invariant under translation and rotation, and cannot handle large-scale deformations well~\cite{gao2017sparse}. Gao \etal~\cite{gao2016efficient} proposed to use the rotation difference on each directed edge to define the deformation. This representation is called RIMD (rotation-invariant mesh difference), which is translation and rotation invariant. RIMD is suitable for mesh interpolation and extrapolation, but reconstructing vertex coordinates from RIMD  requires to solve a very complicated optimization problem. The RIMD feature encodes a plausible deformation space. With the RIMD feature, Tan \etal~\cite{tan2017variational} designed a fully connected mesh variational autoencoder network to extract latent deformation embedding. However, it does not provide disentangled shape and pose latent embeddings for 3D human modeling.

Gao \etal~\cite{gao2017sparse} further proposed another representation called ACAP (as-consistent-as-possible) feature, which allows more efficient reconstruction and derivative computations. Using the ACAP feature, Tan \etal~\cite{tan2017mesh} proposed a convolutional autoencoder to extract localized deformation components from mesh data sets. Gao \etal~\cite{gaoautomatic} also used the ACAP feature to achieve an automatic unpaired shape deformation transfer between two sets of meshes. Wu \etal~\cite{wu2018alive} used a simplified ACAP representation to model caricature face geometry. 

\begin{figure}
\begin{center}
\includegraphics[width=\linewidth]{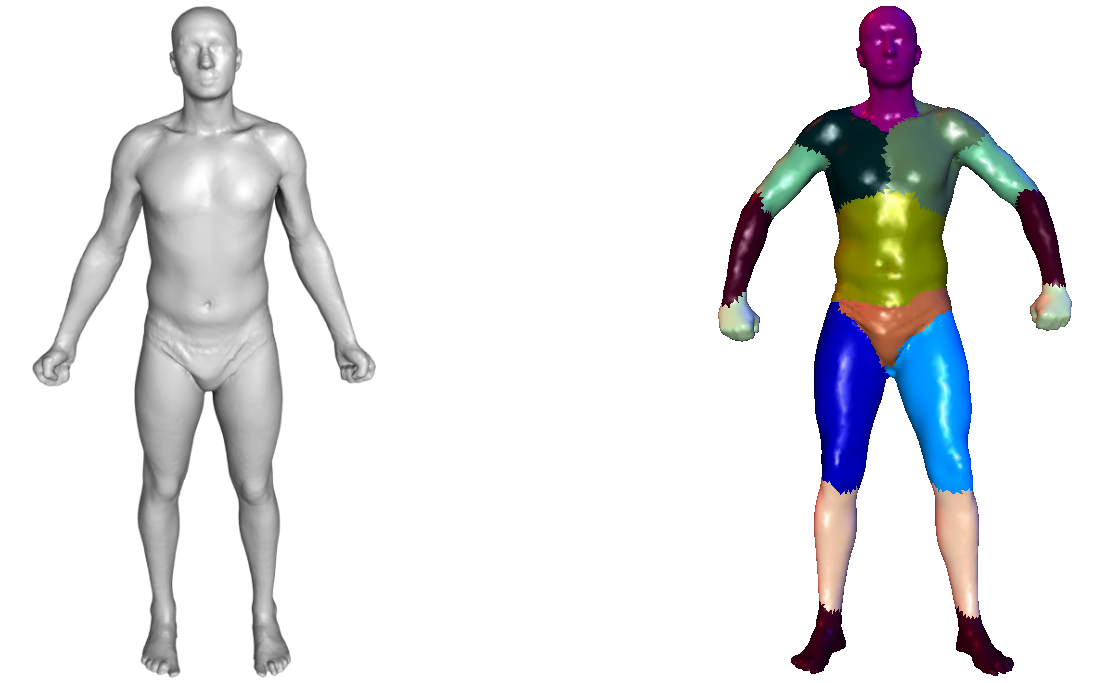}
\end{center}
   \caption{Left: our reference mesh. Right: anatomical human body parts.}
\label{fig:ref_parts}
\end{figure}

\section{Deformation Representation}
\label{sect:acap2m}
This section presents our shape features that are used to represent the human body. 
We adopt a hierarchical architecture to represent and reconstruct the body shape. In particular, we propose a coarse-level shape feature based on anatomical body components and the ACAP feature to represent the human body shape with a general pose.

\textbf{ACAP Feature.} Assume that a mesh dataset consists of $N$ meshes with the same connectivity. We choose one mesh as the reference and the other meshes are considered to be deformed from the reference. We denote the $i$-th vertex coordinates of the reference mesh and the target mesh by $\mathbf{q}_i \in{\mathbb{R}^3}$ and $\mathbf{p}_i \in{\mathbb{R}^3}$, respectively. The deformation at vertex $\mathbf{p}_{i}$ is described locally by an affine transform matrix $\mathbf{T}_{i}\in \mathbb{R}^{3\times3}$ that maps the one-ring neighbor of $\mathbf{q}_{i}$ in the reference mesh to its corresponding vertex on target mesh. The matrix is computed by minimizing
\begin{equation}
\label{equ:local_deform}
\mathop{\arg\min}_{\mathbf{T}_{i}} \sum_{j\in \mathcal{N}(i)} c_{ij}\|(\mathbf{p}_{i} - \mathbf{p}_{j})-\mathbf{T}_{i}(\mathbf{q}_{i} - \mathbf{q}_{j})\|_2^2
\end{equation}
where $c_{ij}$ is the cotangent weight and $\mathcal{N}(i)$ is the index set of one ring neighbor of the $i$-th vertex. Using polar decomposition, $\mathbf{T}_{i} = \mathbf{R}_{i}\mathbf{S}_{i}$, the deformation matrix $\mathbf{T}_{i}$ is decomposed into a rigid component represented by a rotation matrix $\mathbf{R}_{i}$ and a non-rigid component represented by a real symmetry matrix $\mathbf{S}_{i}$. Following~\cite{gao2017sparse}, the rotation matrix $\mathbf{R}_{i}$ can be further represented by a vector $\mathbf{r}_i \in {\mathbb{R}^3}$, and the symmetric matrix $\mathbf{S}_i$ can be represented by a vector $\mathbf{s}_i \in{\mathbb{R}^6}$. To process the ambiguity of axis-angle representation of rotation matrix, Gao \etal~\cite{gao2017sparse} proposed an integer programming approach to solve for optimal $\mathbf{r}_i$ globally and make all $\mathbf{r}_i$ as consistent as possible. Interested readers can refer to~\cite{gao2017sparse} for details. Once $\mathbf{r}_{i}$ and $\mathbf{s}_{i}$ are available, we concatenate all  $[{\mathbf{r}_{i},\mathbf{s}_{i}}]$ together to form the ACAP feature vector $\mathbf{f}\in{\mathbb{R}^{9|\mathcal{V}|}}$ for the target mesh, where $\mathcal{V}$ represents the entire set of mesh vertices. In this way, we convert the target mesh into its ACAP feature representation. As shown in~\cite{gao2017sparse}, by eliminating the ambiguity of axis-angle representation globally,  ACAP feature demonstrates excellent linear interpolation property. Thus ACAP is a good linear space mapping of 3D shape collections with the same connectivity.

\textbf{Coarse Level Deformation Feature.}
 A human body is composed of some anatomical components, and the deformation of a component can be viewed as the main deformation for each vertex belonging to the component. According to the segmentation of~\cite{anguelov2005scape}, we partition a human body into 16 anatomical parts as shown in Fig.~\ref{fig:ref_parts}. We denote by $\mathcal{V}_k$  the set of mesh vertices belonging to the $k$-th part. Similar to Eq.~\eqref{equ:local_deform}, we compute its deformation $\mathbf{T}_{\mathcal{V}_k}$:
\begin{equation}
\label{equ:part_deform}
\mathop{\arg\min}_{\mathbf{T}_{\mathcal{V}_k}} \sum_{i\in \mathcal{V}_k} \|(\mathbf{p}_{i} - \bar{\mathbf{p}}_{\mathcal{V}_k})-\mathbf{T}_{\mathcal{V}_k}(\mathbf{q}_{i} - \bar{\mathbf{q}}_{\mathcal{V}_k})\|_2^2,
\end{equation}
where $\bar{\mathbf{p}}_{\mathcal{V}_k}$ is the mean position of the target mesh's $k$-th part. Similarly, we can represent $\mathbf{T}_{\mathcal{V}_k}$ using $\mathbf{r}_{\mathcal{V}_k} \in {\mathbb{R}^3}$ and $\mathbf{s}_{\mathcal{V}_k} \in {\mathbb{R}^6}$. While axis-angle vector represents the same rotation for $2\pi$ cycle on radian values, which causes ambiguity for $\mathbf{r}_{\mathcal{V}_k}$,  the ACAP feature has eliminated the ambiguities for all $\mathbf{r}_i$, $i\in{\mathcal{V}_k}$. This means that all $\mathbf{r}_i$ have consistent radian values. Therefore, we choose the specific $\mathbf{r}_{\mathcal{V}_k}$ that is closest to the mean of all $\mathbf{r}_i$ of the $k$-th part. Specifically, we modify $\mathbf{r}_{\mathcal{V}_k}$ into
\begin{equation}
\mathbf{r}_{\mathcal{V}_k}=\mathbf{u}_{\mathcal{V}_k}(\theta_{\mathcal{V}_k}+2\pi m),
\end{equation}
where $\theta_{\mathcal{V}_k}$ and $\mathbf{u}_{\mathcal{V}_k}$ are the length and the normalized vector of the initial $\mathbf{r}_{\mathcal{V}_k}$, respectively, and $m$ is computed by solving the following optimization problem
\begin{equation}
\label{equ:part_modify}
\begin{aligned}
m=\mathop{\arg\min}_{j}\|\mathbf{u}_{\mathcal{V}_k}(\theta_{\mathcal{V}_k}+2\pi j ) - \frac{1}{|\mathcal{V}_k|}\sum_{i\in \mathcal{V}_k} \mathbf{r}_i \|_2^2.
\end{aligned}
\end{equation}

\begin{figure}
\begin{center}
\includegraphics[width=\linewidth]{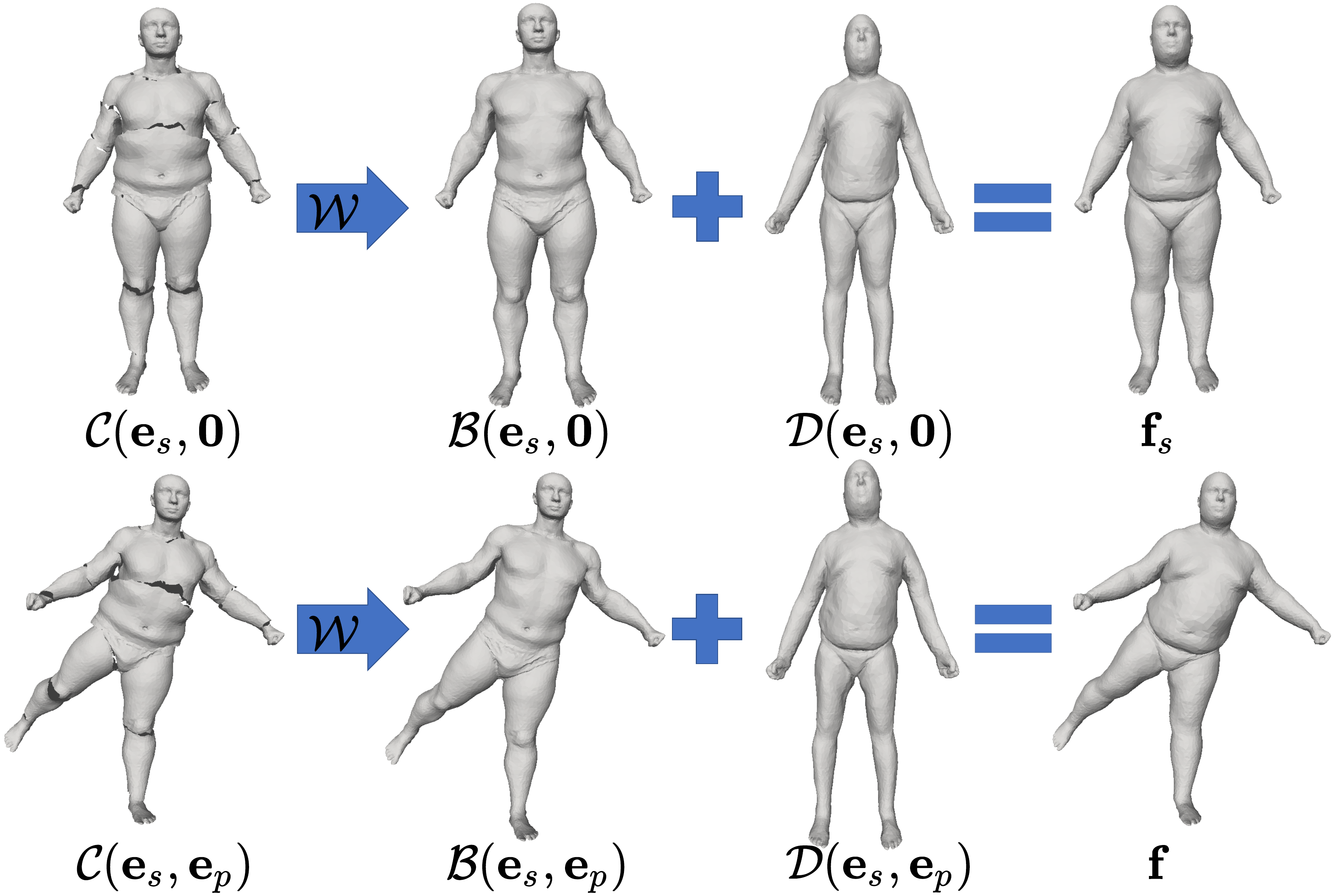}
\end{center}
   \caption{Visualizations of an example shape $\mathbf{f}$. The two rows represent identity and posture decomposition, respectively. The first three columns show coarse shapes $\{\mathbf{g}_s,\mathbf{g}\}$, base shapes $\{\mathbf{b}_s,\mathbf{b}\}$ and difference shapes $\{\mathbf{d}_s,\mathbf{d}\}$ in the hierarchical reconstruction pipeline.}
\label{fig:overview}
\end{figure}

Once $\mathbf{r}_{\mathcal{V}_k}$ and $\mathbf{s}_{\mathcal{V}_k}$ are found for all parts, we concatenate all $[\mathbf{r}_{\mathcal{V}_k},\mathbf{s}_{\mathcal{V}_k}]$ together to form the coarse-level feature $\mathbf{g}\in{\mathbb{R}^{144}}$. Each $[\mathbf{r}_{\mathcal{V}_k},\mathbf{s}_{\mathcal{V}_k}]$ encodes the optimal affine transformation of the $k$-th part relative to the reference part. In the first column of Fig.~\ref{fig:overview}, we show a group of coarse level deformation shapes of target meshes.

\textbf{ACAP to Mesh.} Converting a given ACAP feature vector $\mathbf{f}\in{\mathbb{R}^{9|\mathcal{V}|}}$ to the target mesh is easy. In particular, we directly reconstruct $\mathbf{T}_{i}$ from $[\mathbf{r}_{i},\mathbf{s}_{i}]$~\cite{gao2017sparse}. The vertex coordinates $\mathbf{p}_i$ of the target mesh can be obtained by solving
\begin{equation}
\label{equ:local_deform_vertices}
\mathop{\arg\min}_{\{\mathbf{p}_i\}} \sum_{j\in \mathcal{N}(i)} c_{ij}\|(\mathbf{p}_{i} - \mathbf{p}_{j})-\mathbf{T}_{i}(\mathbf{q}_{i} - \mathbf{q}_{j})\|_2^2
\end{equation}
which is equivalent to the following system of linear equations:
\begin{equation}
\label{equ:recons}
    2\sum_{j\in{N(i)}}c_{ij}\mathbf{e}_{ij}=\sum_{j\in{\mathcal{N}(i)}}c_{ij}(\mathbf{T}_i+\mathbf{T}_j)(\mathbf{q}_j - \mathbf{q}_i),
\end{equation}
where $\mathbf{e}_{ij}=\mathbf{p}_j-\mathbf{p}_i$. Note that Eq.~\eqref{equ:recons} is translation-independent. Thus we need to specify the position of one vertex. Then the amended linear system can be rewritten as $\mathbf{A}\mathbf{p}=\mathbf{b}$ where $\mathbf{A}$ is a fixed and sparse coefficient matrix, for which a pre-decomposing operation can be executed to save the computation time.

\textbf{Scaling Deformation Feature.} Following the strategy of Tan \etal~\cite{tan2017variational}, we rescale each dimension of $\mathbf{f}$ and $\mathbf{g}$ to $[-0.95,0.95]$ independently. This strategy normalizes each dimension of the features and reduces learning difficulty of reconstructing deformation features $\mathbf{f}$ and $\mathbf{g}$.

\section{Overview}
\label{sect:overview}
This section gives a detailed description of our proposed representation for 3D human body. We adopt the ACAP feature to represent human body shape considering its linear space mapping for large scale deformation. With the ACAP feature, we can use addition operations to represent the composition of non-rigid deformations.

Our proposed human body representation is motivated by the following two factors: semantics and precision. For semantics, an identity and pose disentangled body representation is required for many human body related applications. Therefore, for an ACAP feature $\mathbf{f}$ of a human body, we denote its latent parameters by a set of disentangled parameters $\{\mathbf{e}_s, \mathbf{e}_p\}$, where $\mathbf{e}_s$ and $\mathbf{e}_p$ control the shape variations determined by identity and posture, respectively. We define the neutral pose shape feature $\mathbf{f}_s$ of $\mathbf{f}$ as the ACAP feature decoded from $\{\mathbf{e}_s,\mathbf{0}\}$. The last column of Fig.~\ref{fig:overview} shows a posed human body and its corresponding neutral body. In this paper, the latent representation denotes a compressed representation of the original shape model, which is the only information the decoder is allowed to use to reconstruct the input shape model as faithfully as possible.

To improve the representation accuracy, we adopt a hierarchical strategy. Specific to the human body, a natural idea is to utilize the deformation of anatomical components as the bridge to the final shape. From Section~\ref{sect:acap2m}, we know that the deformation of body components is encoded by the coarse level deformation feature $\mathbf{g}\in{\mathbb{R}^{9\times16}}$ of $\mathbf{f}\in{\mathbb{R}^{9|\mathcal{V}|}}$. We use $\mathcal{C}(\mathbf{e}_s,\mathbf{e}_p)$ to represent the mapping from the latent parameters to $\mathbf{g}$, and denote the coarse feature of $\mathbf{f}_s$ by $\mathbf{g}_s$. The deformation of components encoded by $\mathbf{g}$ has much lower dimensions than $\mathbf{f}$, and each vertex feature of $\mathbf{f}$ encompasses similar base deformation determined by related components. Therefore, based on the articulate structure, we model a base part $\mathbf{b} = \mathcal{B}(\mathbf{e}_s, \mathbf{e}_p)$ of $\mathbf{f}$ with $\mathcal{W}(\mathcal{C}(\mathbf{e}_s, \mathbf{e}_p))$, where $\mathcal{W}$ is a linear blend skinning operation that recovers the deformation of each vertex on $\mathbf{b}$ by linearly blending the deformations of related components on $\mathbf{g}$. Similarly, we use $\mathbf{b}_s$ to represent the neutral counterpart of $\mathbf{b}$. A group of coarse features and base features is visualized in the first two columns of Fig.~\ref{fig:overview}.

Considering that base feature $\mathbf{b}$ only encodes the optimal affine transformation relative to the reference mesh based on anatomical components, which does not include the fine-scale deformations caused by identities, soft tissues movement and different postures, we introduce difference features $\mathbf{d} = \mathcal{D}(\mathbf{e}_s,\mathbf{e}_p)$ and $\mathbf{d}_s = \mathcal{D}(\mathbf{e}_s,\mathbf{0})$ to recover $\mathbf{f}$ and $\mathbf{f}_s$ better. Our final proposed human body representation can be expressed as:
\begin{equation}
\label{equ:architecture_d}
\begin{aligned}
\mathbf{f}=\mathcal{W}(\mathcal{C}(\mathbf{e}_s,\mathbf{e}_p))+\mathcal{D}(\mathbf{e}_s,\mathbf{e}_p)
\end{aligned}
\end{equation}
and we can further represent $\mathcal{C}$ and $\mathcal{D}$ as:
\begin{equation}
\begin{aligned}
\mathcal{C}(\mathbf{e}_s,\mathbf{e}_p)=\mathcal{T}_c(\mathcal{C}_s(\mathbf{e}_s)+\mathcal{C}_p(\mathbf{e}_p)), \\
\mathcal{D}(\mathbf{e}_s,\mathbf{e}_p)=\mathcal{T}_d(\mathcal{D}_s(\mathbf{e}_s)+\mathcal{D}_p(\mathbf{e}_p)).
\end{aligned}
\end{equation}
$\mathcal{C}(\mathbf{e}_s,\mathbf{e}_p)$ aims to reconstruct coarse level deformation feature $\mathbf{g}$ by summing two independent parts $\mathcal{C}_s(\mathbf{e}_s)$ and $\mathcal{C}_p(\mathbf{e}_p)$ and then applying a mapping $\mathcal{T}_c$, which is introduced to enhance the non-linearity of the representation and thus improve its expression ability. For difference feature $\mathcal{D}(\mathbf{e}_s,\mathbf{e}_p)$, we follow the same design. As shown in the first row of Fig.~\ref{fig:overview}, we can get all neutral pose counterpart features $\mathbf{g}_s,\mathbf{b}_s$ and $\mathbf{d}_s$ of all corresponding features by setting $\mathbf{e}_p$ to $\mathbf{0}$. For body representation in Eq.~\eqref{equ:architecture_d}, each mapping can be implemented with MLP (multilayer perceptron) with arbitrary complexity. In this way, an end-to-end neural network can be integrated with this representation.

\begin{figure*}[t!]
\begin{center}
\includegraphics[width=\linewidth]{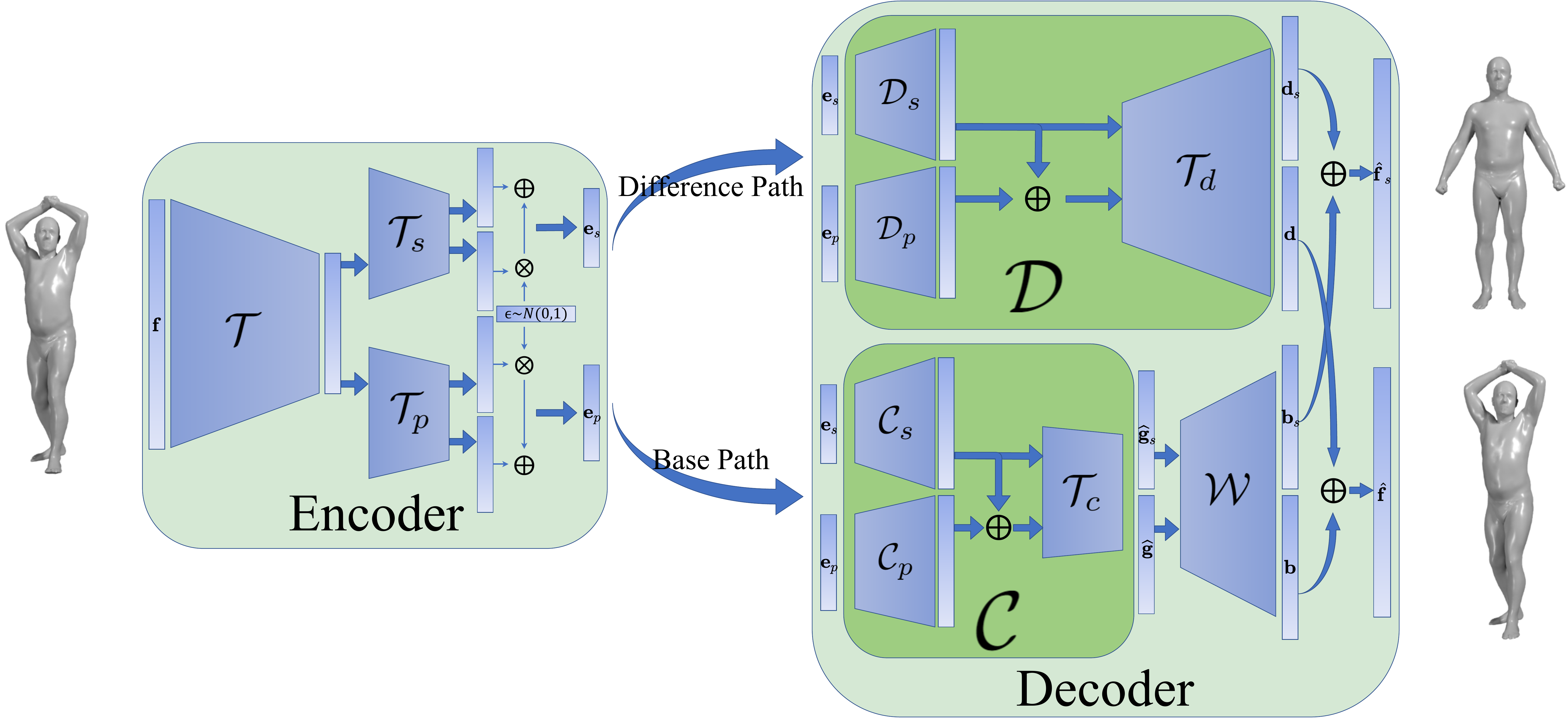}
\end{center}
    \caption{The architecture of our proposed embedding learning network. The Encoder encodes ACAP feature $\mathbf{f}$ into shape and pose latent pair $\{\mathbf{e}_s, \mathbf{e}_p\}$. The Decoder has two decoding paths to generate base and difference features, respectively. The base feature captures large scale deformation determined by anatomical human body parts, and the difference feature describes the relatively small scale difference caused by different pose, identity, and soft tissue movement. The final reconstructed feature $\hat{\mathbf{f}}$ can be recovered by summing the base feature $\mathbf{b}$ and difference feature $\mathbf{d}$. By setting pose latent code $\mathbf{e}_p$ to be $\mathbf{0}$, we reconstruct the corresponding neutral body shape $\hat{\mathbf{f}}_s$.}
\label{fig:architecture}
\end{figure*}

In the next few sections, we will give the implementation details of our proposed human body representation. In particular, we first present our neural network architecture and loss function design in Section~\ref{sect:learning}. Then we give the construction of our body shape dataset in Section~\ref{sec:dataset}, and we show how to use the proposed learned embedding in Section~\ref{sect:use}. Finally, the detailed experimental evaluations are reported in Section~\ref{sect:experiment}.

\section{Embedding Learning} \label{sect:learning}
\subsection{Network Architecture}
\label{sect:architecture}
In this work, our goal is to learn a disentangled human body representation with a hierarchical reconstruction pipeline. We define the coarse shape, the base shape, the difference shape and the body shape as $\mathbf{g}, \mathbf{b}, \mathbf{d}, \mathbf{f}$, respectively. To learn disentangled and hierarchical representation, we need large scale training data with ground truth $\{\mathbf{f},\mathbf{g},\mathbf{f}_s,\mathbf{g}_s\}$ to supervise our embedding learning.

We use a VAE like architecture in our end-to-end representation learning.  Fig.\ref{fig:architecture} shows the proposed architecture. For the encoder, we first feed $\mathbf{f} \in{\mathbb{R}^{9|\mathcal{V}|}}$ into a shared MLP (multilayer perceptron) $\mathcal{T}$ to generate a 400 dimension hidden feature. Then we use the standard VAE~\cite{kingma2013auto} encoder structure $\{\mathcal{T}_s,\mathcal{T}_p\}$ to generate the shape and pose latent representations $\{\mathbf{e}_s,\mathbf{e}_p\}$ separately. Specially, $\mathcal{T}$ is composed of two fully connected layers with $tanh$ as the activation function. $\{\mathcal{T}_s,\mathcal{T}_p\}$ have similar structure and they use a fully connected layer without activation to output the mean values and another fully connected layer with $2\times sigmoid$ activation to output the standard deviation. We set the shape embedding $\mathbf{e}_s$ to $50$ dimensions and the pose embedding $\mathbf{e}_p$ to $72$ dimensions, \ie,  $\mathbf{e}_s\in{\mathbb{R}^{50}}$ and $\mathbf{e}_p\in{\mathbb{R}^{72}}$, to roughly match the dimensions of the shape and pose parameters in SMPL~\cite{loper2015smpl}.

Our decoder follows the design of Eq.~\eqref{equ:architecture_d}. There are two paths called base path and difference path. Each path takes $\{\mathbf{e}_s,\mathbf{e}_p\}$ as input, and corresponds to $\mathcal{W}(\mathcal{C}(\mathbf{e}_s,\mathbf{e}_p))$ and $\mathcal{D}(\mathbf{e}_s,\mathbf{e}_p)$ in Eq.~\eqref{equ:architecture_d}, respectively. The decoder outputs $\hat{\mathbf{f}}$ by summing reconstructed base feature $\mathbf{b}$ and difference feature $\mathbf{d}$ of the two paths and produces $\hat{\mathbf{g}}$ with $\mathcal{C}(\mathbf{e}_s,\mathbf{e}_p)$, and $\{\hat{\mathbf{f}},\hat{\mathbf{g}}\}$ aims to reconstruct $\{\mathbf{f},\mathbf{g}\}$. Meanwhile, the decoder outputs $\{\hat{\mathbf{f}}_s,\hat{\mathbf{g}}_s\}$ by another calculation with $\{\mathbf{e}_s,\mathbf{0}\}$ as inputs, where $\{\hat{\mathbf{f}}_s,\hat{\mathbf{g}}_s\}$ aim to reconstruct $\{\mathbf{f}_s,\mathbf{g}_s\}$. The detailed structure of the decoder is given in the Appendix.

The learnable skinning layer $\mathcal{W}$ is introduced to construct base feature $\mathbf{b}\in{\mathbb{R}^{9|\mathcal{V}|}}$ from coarse level feature $\mathbf{g}\in{\mathbb{R}^{144}}$. The skinning method has showed its ability for human body modeling based on Euclidean coordinates~\cite{loper2015smpl}. Our learnable skinning layer exploits this method in the feature space. Particularly, we use a learnable sparse matrix $\mathbf{W}\in{\mathbb{R}^{|\mathcal{V}|\times16}}$ to transform coarse level feature $\mathbf{g}\in{\mathbb{R}^{16\times9}}$ to base feature $\mathbf{b}\in{\mathbb{R}^{|\mathcal{V}|\times9}}$, \ie,

\begin{equation}
\label{equ:skinning}
\begin{aligned}
\mathbf{b}&=\mathbf{W}\mathbf{g}, \\
s.t. \mathbf{W}_{ij}\geq 0 \quad &\sum_{j=1}^{16} \mathbf{W}_{ij} &= 1,
\end{aligned}
\end{equation}
where each row of $\mathbf{b}$ is a convex combination of rows of coarse-level feature $\mathbf{g}$. Moreover, we constrain $\mathbf{W}_i$ to be non-zero only on the nearby parts of the $i$-th vertex to avoid an overfitting and non-smoothing solution.

\subsection{Loss Function}
\label{sec:train}
We use $\ell_{1}$ error for the feature reconstruction:
\begin{equation}
    \begin{aligned}
E_{L1_1}=\frac{1}{9|\mathcal{V}|} \|\hat{\mathbf{f}}_{s} - \mathbf{f}_{s}\|_{1},\quad
E_{L1_2}=\frac{1}{9|\mathcal{V}|} \|\hat{\mathbf{f}} - \mathbf{f}\|_{1}.
    \end{aligned}
\end{equation}
Similarly, for coarse-level feature reconstruction, we define
\begin{equation}
    \begin{aligned}
        E_{L1_{c_1}}=\frac{1}{9\times16} \|\hat{\mathbf{g}}_{s}-\mathbf{g}_{s}\|_{1}\quad,
        E_{L1_{c_2}}=\frac{1}{9\times16} \|\hat{\mathbf{g}} - \mathbf{g}\|_{1}.
    \end{aligned}
\end{equation}

For the shape and pose embedding, since we use VAE as the encoder, KL divergence losses are needed to regularize the distribution of latent parameters:
\begin{equation}
\begin{aligned}
    E_{sKL}=D_{KL}(q(\mathbf{e}_{s}|\mathbf{f})\|p(\mathbf{e}_{s})), \\
    E_{pKL}=D_{KL}(q(\mathbf{e}_{p}|\mathbf{f})\|p(\mathbf{e}_{p})),
\end{aligned}
\end{equation}
where $q(\mathbf{e}|\mathbf{f})$ is the posterior probability, $p(\mathbf{e})$ is the prior multivariate normal distribution, and $D_{KL}$ is the KL divergence formulation. See~\cite{kingma2013auto} for more details of the KL divergence formulation. The total loss is given in the following form:

\begin{equation}
\label{eq:loss}
\begin{aligned}
    Loss&=\lambda_{s}E_{sKL}+\lambda_{r_1}E_{L1_1}+\lambda_{r_{c_1}}E_{L1_{c_1}}\\
    &+\lambda_{p}E_{pKL}+\lambda_{r_2}E_{L1_2}+\lambda_{r_{c_2}}E_{L1_{c_2}}.
\end{aligned}
\end{equation}
The configuration details of all related hyperparameters and the choice of loss function are given in the Appendix.

\section{Constructing Training Data}
\label{sec:dataset}
To facilitate data-driven 3D human body analysis, we need to have a large number of 3D human mesh models. Thus, we collect data from several publicly available datasets. In particular, SCAPE~\cite{anguelov2005scape} and FAUST~\cite{bogo2014faust} provide meshes of several subjects with different poses. Hasler \etal~\cite{hasler2009statistical} provide 520 body meshes for about 100 subjects with relatively low resolution. MANO~\cite{MANO:SIGGRAPHASIA:2017} collects the body and hand shapes of several people. Dyna~\cite{pons2015dyna} and DFaust~\cite{bogo2017dynamic} release the alignments of several subjects' movement scan sequences. For the rest-pose body data set, CAESAR database~\cite{robinette1999caesar} is the largest commercially available dataset that contains 3D scans of over 4500 American and European subjects in a standard pose. Yang \etal~\cite{yang2014semantic} convert a large part of the CAESAR dataset to the same connectivity with the SCAPE dataset. All these datasets have very different connectivity structures and different poses for each identity.

Our proposed embedding learning network has two main requirements for the training data. First, the connectivity of the whole dataset must be the same to facilitate the ACAP feature computation. Second, to disentangle human body variations into shape and pose latent embeddings, we need to define a neutral pose as the specific pose that represents the body variations only caused by identity, \ie, intrinsic factors among individuals. In other words, we need to construct a neutral pose mesh for each identity in our dataset.

For the first requirement, we need to convert our collected public datasets, like FAUST~\cite{bogo2014faust}, SCAPE~\cite{anguelov2005scape} and Hasler \etal~\cite{hasler2009statistical} into the same connectivity. Considering vertex density and data amount, we modify the connectivity shared by SCAPE~\cite{anguelov2005scape} and SPRING~\cite{yang2014semantic} to eliminate several non-manifold faces and treat this connectivity as the standard one. Specifically, we set the mesh graph structure with $|\mathcal{V}|=12500$ vertices and 24495 faces, which is much denser than SMPL~\cite{loper2015smpl} that has 6890 vertices. We choose one mesh of SCAPE~\cite{anguelov2005scape} as the reference mesh, as shown in Fig.~\ref{fig:ref_parts}, for the ACAP feature computation.

For the second requirement, SPRING~\cite{yang2014semantic} is a dataset with a consistent and simple pose, which can be regarded as our neutral pose.
\begin{figure*}[t]

\begin{center}
\includegraphics[width=\linewidth]{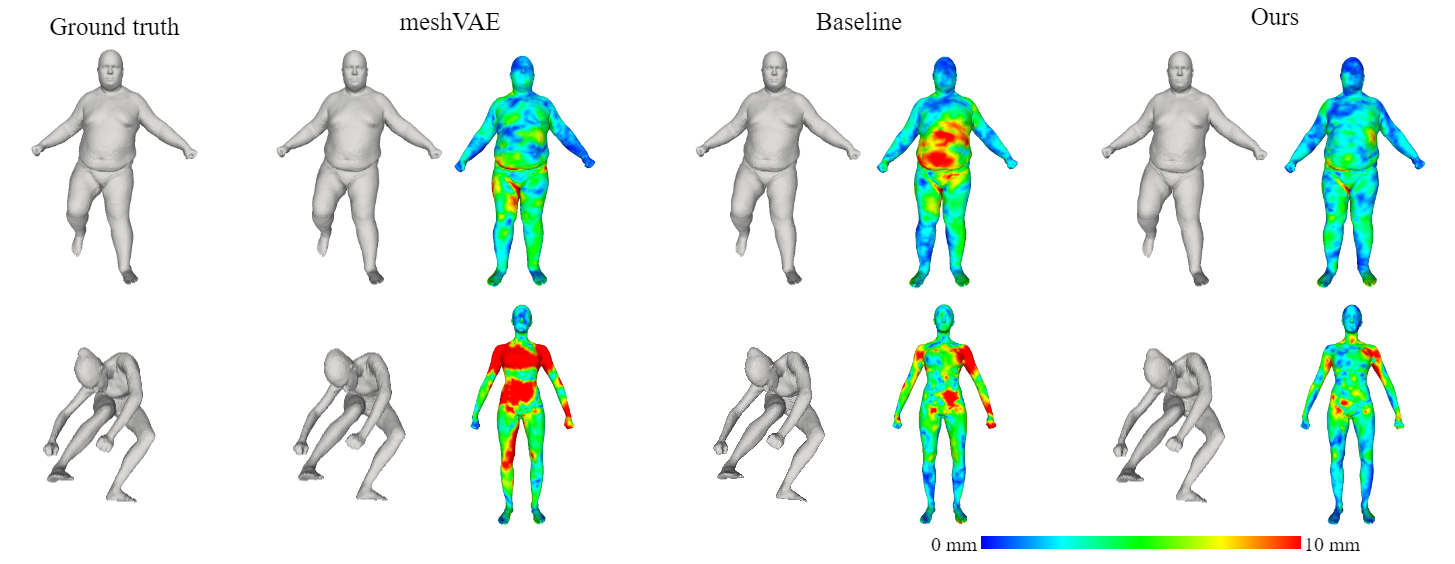}
\end{center}
   \caption{Comparison of three methods in reconstructing two samples from test pose date, which have different shapes and poses. For each method, the left column shows the reconstruction results and the right column shows the MED error maps on a neutral model.}
\label{fig:test_rec}
\end{figure*}

\textbf{Connectivity Conversion.}
We formulate our connectivity conversion problem as a non-rigid registration problem from the reference connectivity to a mesh in a target connectivity dataset. We adopt the data-driven non-rigid deformation method of Gao \etal~\cite{gao2017sparse} to solve our problem. First, we define the prior human body deformation space by a base of ACAP features. We use 70 pose meshes of SCAPE~\cite{anguelov2005scape} to cover the pose variations, and choose 70 shape meshes of different individuals from SPRING~\cite{yang2014semantic} to cover the shape variations. With the computed 140 ACAP features (see Section~\ref{sect:acap2m}), we get a matrix $\mathbf{F}\in\mathbb{R}^{9|\mathcal{V}|\times 140}$. Then, we extract a sparse base $\mathbf{C}\in\mathbb{R}^{9|\mathcal{V}|\times K}$ from $\mathbf{F}$, by using the sparse dictionary learning method~\cite{neumann2013sparse}. Unlike~\cite{gao2017sparse}, we extract the sparse base based on human body parts instead of automatically selecting the basis deformation center. See Fig.~\ref{fig:ref_parts} for the segmentation of human body parts.
In this way, we can now use a vector $\mathbf{w}\in{\mathbb{R}^K}$ to obtain an ACAP feature: $\mathbf{f}(\mathbf{w})=\mathbf{C}\mathbf{w}.$

Second, we manually mark a set of corresponding vertices between the reference and the target connectivity, denoted as $\{i, l(i)\},i\in{\mathcal{L}}$, where $\mathcal{L}$ is the index set of markers on our reference connectivity and $l(i)$ represents the index of the corresponding marker on the target connectivity.

Finally, we formulate our connectivity conversion problem as:
\begin{equation}
    \label{equ:regist}
    \mathop{\arg\min}_{\mathbf{R},\mathbf{t},\mathbf{p}_{i},\mathbf{w}} E_{\textrm{prior}} + \lambda_1 E_{\textrm{icp}}+\lambda_2 E_{\textrm{lan}} + \lambda_3 \|\mathbf{w}\|_1
\end{equation}
$$E_{\textrm{prior}}=\sum_{i}\sum_{j\in \mathcal{N}(i)}c_{ij}\|(\mathbf{p}_i-\mathbf{p}_j)-\mathbf{T}_i(\mathbf{w})(\mathbf{q}_{i}-\mathbf{q}_{j})\|_2^2$$
$$E_{\textrm{icp}}=\sum_{i}\|\mathbf{n}_{l(i)}^T(\mathbf{R}\mathbf{p}_i+\mathbf{t}-\mathbf{v}_{l(i)})\|$$
    $$E_{\textrm{lan}}=\sum_{i\in{\mathcal{L}}}\|\mathbf{R}\mathbf{p}_i+\mathbf{t} - \mathbf{v}_{l(i)}\|_2^2$$
where $\mathbf{R}$ and $\mathbf{t}$ represent the rotation and translation of the global rigid transformation, $E_{\textrm{icp}}$ is the point-to-plane ICP energy, $\mathbf{n}_{l(i)}$ is the normal of vertex $\mathbf{v}_{l(i)}$ on the target mesh, $\mathbf{p}_{i}$ is a vertex to be optimized on the reference mesh connectivity, $\mathbf{q}_{i}$ is a vertex of the reference mesh, and $\mathbf{E}_{\textrm{lan}}$ is for sparse landmark constraints. $E_{\textrm{prior}}$ is the formulation from Gao \etal~\cite{gao2017sparse}, which uses the extracted sparse deformation base $\mathbf{C}$ to generate transformation $\mathbf{T}_{i}(\mathbf{w})$ so as to constrain the movements of $\mathbf{p}_i$. By default, we set $\lambda_1$, $\lambda_2$ and $\lambda_3$ to 5.0, 1.0 and 0.3, respectively.

By using this connectivity conversion method, we convert 916 meshes from Dyna~\cite{pons2015dyna}, all 100 meshes of FAUST~\cite{bogo2014faust}, 517 meshes of Hasler \etal~\cite{hasler2009statistical} and 852 meshes of MANO~\cite{MANO:SIGGRAPHASIA:2017} to the standard connectivity and align the converted meshes to the reference mesh.

\textbf{Neutral Pose Construction.}
We compute the average shape of SPRING~\cite{yang2014semantic} as the target neutral pose. For each subject, we choose the posture mesh with the smallest rigid transformation to the target as the reference mesh, and apply ARAP (as rigid as possible) deformation~\cite{sorkine2007rigid} to deform the reference mesh to the target neutral pose. Specifically, we manually label several landmark pairs for both meshes on arms, forearms, legs, spine, etc. Then we use the deviation of orientations determined by each landmark pair on both meshes as soft constraints to deform the posture mesh to the neutral pose with ARAP deformation. In this way, we generate another 135 neutral meshes.

Finally, with the method described above, we obtain 2385 converted pose meshes plus another 70 from SCAPE~\cite{anguelov2005scape}, and 135 deformed neutral meshes plus 3048 from SPRING~\cite{yang2014semantic}. We compute their ACAP features $\mathbf{f}$ and corresponding coarse level features $\mathbf{g}$ using the method described in Section~\ref{sect:acap2m}. After removing a few bad results, we eventually get 5594 pair features. We choose the corresponding neutral features $\{\mathbf{f}_s,\mathbf{g}_{s}\}$ for every pair $\{\mathbf{f},\mathbf{g}\}$, and construct the final dataset. Then, we randomly choose 160 neutral meshes, and 160 pose meshes as testing data. And the rest are used as training data. Table~\ref{tab:dataset} shows the numbers of meshes used from each dataset and in our constructed dataset.

\begin{figure}
\begin{center}
\includegraphics[width=\linewidth]{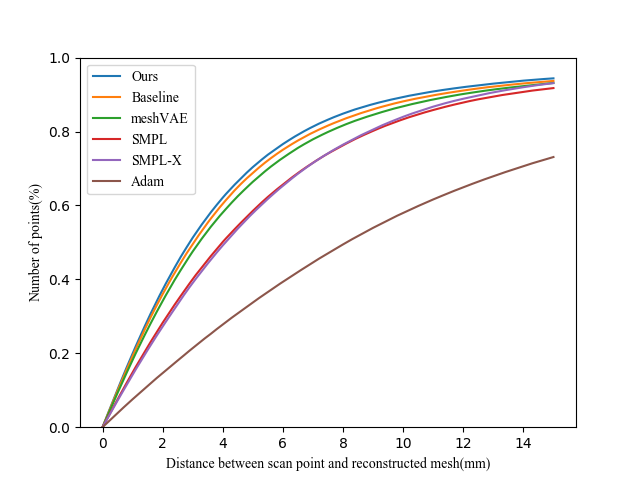}
\end{center}
   \caption{Cumulative Errors Distribution (CED) curves for our shape scan dataset.}
\label{fig:Apose_curve}
\end{figure}

\begin{figure*}[t]
\begin{center}
\includegraphics[width=\linewidth]{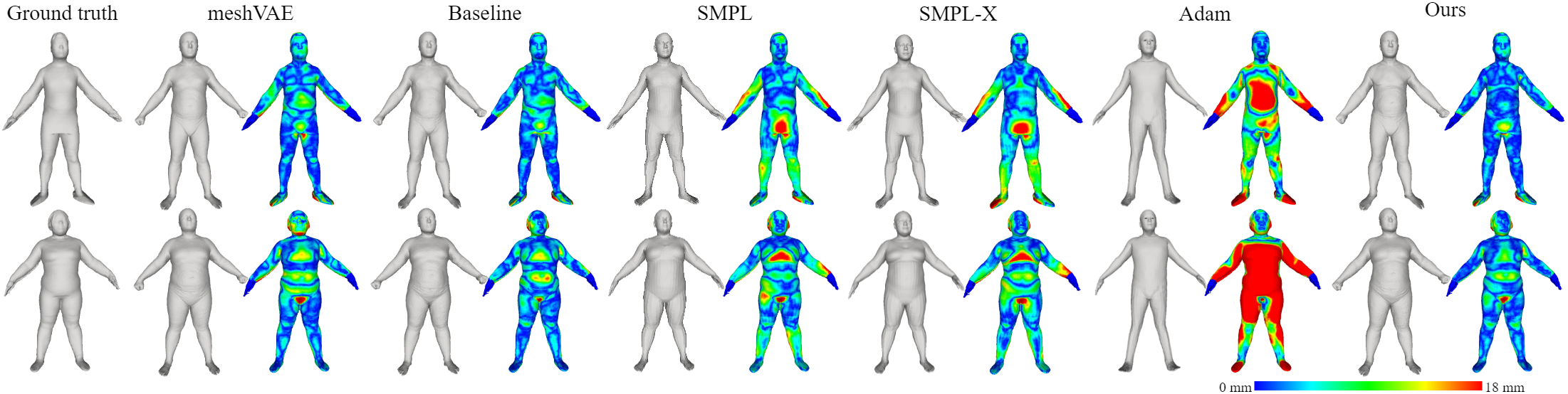}
\end{center}
   \caption{Comparison of the six methods on two samples from the shape scan dataset. The reconstruction results and the PMD error maps on scan point clouds are displayed.}

\label{fig:Apose_errormap}
\end{figure*}

\begin{table}
\begin{center}
\caption{The second column shows the number of our constructed neutral and pose meshes. We also present the number of meshes used from existing datasets.}
\label{tab:dataset}
\vskip -0.5cm
\begin{tabular}{|c|c|c|c|c|c|}
\hline
DataSet &\textbf{Neutral} &SPRING~\cite{yang2014semantic}&SCAPE~\cite{anguelov2005scape}&Hasler \etal~\cite{hasler2009statistical} \\
\hline
number & $\mathbf{3183}$ & 3048 & 70 &  517  \\
\hline
DataSet &\textbf{Pose} &FAUST~\cite{bogo2014faust}&Dyna~\cite{pons2015dyna}& MANO~\cite{MANO:SIGGRAPHASIA:2017}\\
\hline
number & $\mathbf{2411}$ & 99  & 907 & 818   \\
\hline
\end{tabular}
\end{center}
\vspace{-0.1in}
\end{table}

\begin{figure}
\begin{center}
\includegraphics[width=\linewidth]{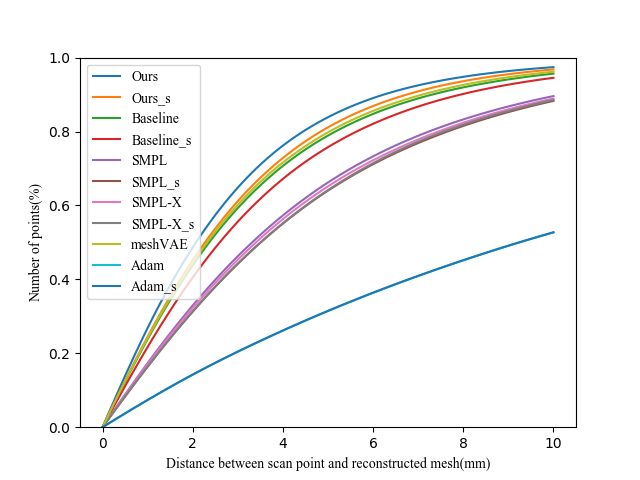}
\end{center}
   \caption{Cumulative Errors Distribution(CED) curves for DFaust~\cite{bogo2017dynamic} scan dataset.}
\label{fig:DFaust_dense}
\end{figure}

\section{Use of the Embedding} \label{sect:use}
Once the embedding learning is done, we only need to keep the trained decoder plus the ACAP-to-mesh converter of Eq.~\eqref{equ:recons}. We denoted this generator by $\mathcal{M}(\mathbf{e}_s, \mathbf{e}_p)$, which takes shape and pose parameters $\{\mathbf{e}_s, \mathbf{e}_p\}$ as input and outputs a mesh in the predefined connectivity. For various online applications such as reconstruction, we just need to optimize the low-dimensional embedding $\mathcal{M}(\mathbf{e}_s, \mathbf{e}_p)$ to fit the input data, which could be the image, video, point cloud, mesh, etc.

Let us use the mesh input as an example. Given a mesh with our pre-determined connectivity, we want to find the optimal $\{\mathbf{e}_s^*, \mathbf{e}_p^*\}$ whose $\mathcal{M}(\mathbf{e}_s^*, \mathbf{e}_p^*)$ best reconstructs the given mesh. Here, we do not use our trained encoder to obtain $\{\mathbf{e}_s, \mathbf{e}_p\}$ since the encoder requires to convert the given mesh into ACAP features, which is complex and time-consuming. Instead, we optimize $\{\mathbf{e}_s, \mathbf{e}_p\}$ directly by only using the decoder:
\begin{equation} \label{equ:decoder}
    \min_{\mathbf{e}_s, \mathbf{e}_p,\mathbf{R},\mathbf{t}} \lambda\sum_i^{|\mathcal{V}|}\|\mathbf{R} {\mathbf{p}}_i(\mathbf{e}_s, \mathbf{e}_p)+\mathbf{t}-\mathbf{q}_i\|_2^2 + \lambda_{\beta}\|\mathbf{e}_s\|_2^2 +\lambda_{\theta}\|\mathbf{e}_p\|_2^2
\end{equation}
where rotation $\mathbf{R}$ and translation $\mathbf{t}$ are the global rigid transformation parameters, ${\mathbf{p}}_i$ is the $i$-th vertex position of the decoded mesh of $\mathcal{M}(\mathbf{e}_s, \mathbf{e}_p)$, and $\mathbf{q}_i$ is the $i$-th vertex of the given mesh. For this optimization with per vertex constraints, we assign $\lambda$ to $1.0\times10^6$, $\lambda_{\beta}$ and $\lambda_{\theta}$ to $1.0$. This model generally takes about 200 iterations to achieve millimeter reconstruction accuracy with Adam optimization.

\begin{table}
\begin{center}
\caption{MED($mm$) for the test dataset consisting of 160 neutral meshes and 160 pose meshes.}
\label{tab:test_recons}
\begin{tabular}{|c|c|c|c|c|}
\hline
Test Dataset & Ours &Baseline  &meshVAE \\
\hline
Neutral (160) & $\mathbf{4.67}$ &4.99  &5.26 \\
Pose (160) & $\mathbf{2.75}$ &3.19 &3.13 \\
\hline
\end{tabular}
\end{center}
\vspace{-0.1in}
\end{table}

\begin{table}
\begin{center}
\caption{Quantitative comparison of different methods on our shape scan dataset. The mean PMD($mm$), the standard deviation, and the valid number of points for testing (without hand part) are given. We also give the errors and the standard deviations (wh) for all points in the last two columns just for reference, although our method does not consider the hand part.}
\label{tab:Apose_table}
\begin{tabular}{|c|c|c|c|c|c|}
\hline
Methods &mean & std & \#points &mean(wh) & std(wh) \\
\hline
Ours & $\mathbf{4.9}$ & $\mathbf{6.8}$ &545263 &6.5 & 10.9 \\
Baseline & 5.2 & 7.5 &543848 &7.0 & 11.9 \\
meshVAE & 5.4 & 7.2 & 544794&6.9 & 11.1 \\
SMPL & 6.4 & 8.5 & 546020&6.4 & 8.2 \\
SMPL-X & 6.1 & 7.2 & 543853&$\mathbf{6.1}$ & $\mathbf{7.1}$ \\
Adam & 12.1 & 13.3 & 547843&11.5 & 12.9 \\
\hline
\end{tabular}
\end{center}
\vspace{-0.1in}
\end{table}

\section{Experiments} \label{sect:experiment}

In this section, we quantitatively evaluate our model's capability for reconstruction and present some qualitative results and potential applications. We set several baseline methods for comparison in different tasks. To show the benefit of our hierarchical reconstruction pipeline, we train a baseline architecture called ``Baseline'' that removed the base path in the decoder. To compare the effect of disentangling shape and pose variations, we train the non-disentangled meshVAE~\cite{tan2017variational} architecture on our dataset. To evaluate the representation ability, we compare our method with the widely used SMPL model~\cite{loper2015smpl} and its variant SMPL-X model~\cite{pavlakos2019expressive}. We also perform a comparison with the Adam body model~\cite{xiang2019monocular}, even though it is mainly designed to estimate body movement rather than body geometry. We integrate the official gender-neutral model code into the PyTorch framework and implement the optimization in the same framework with Adam~\cite{kingma2014adam}.

\textbf{Computation Time}. Our implementation is based on PyTorch. Our mesh decoder $\mathcal{M}(\mathbf{e}_s, \mathbf{e}_p)$ takes about 10ms to map an embedding to a mesh on TITAN Xp GPU.

\subsection{Quantitative Evaluation}
\label{sec:rec_accuracy}
In this section, we evaluate the performance of reconstruction, 3D pose estimation, and alignment on FAUST. For reconstruction, we perform quantitative evaluations on two types of data. The first type of data is from our test dataset, where all the meshes have the same connectivity. We use the mean Euclidean distance of vertices (MED) as the measurement. The second type of data is the general scan data of human bodies.  We compute the distance between each point of scan point cloud and the corresponding reconstructed mesh as the measure. The distance is computed with the AABB tree, and we denote this error measurement by PMD (point-mesh distance). Note that all test point clouds are obtained by scanning the human body with an open hand, while the fists of our template body mesh are closed. Thus it is unfair to include the scan points of hand parts when comparing with SMPL and SMPL-X as they have an open hand model. Therefore when computing the PMD values, we ignore the hand part and mainly focus on the body part. For reference, we also report the errors of all points in related tables.

\begin{figure*}[t]
\begin{center}
\includegraphics[width=\linewidth]{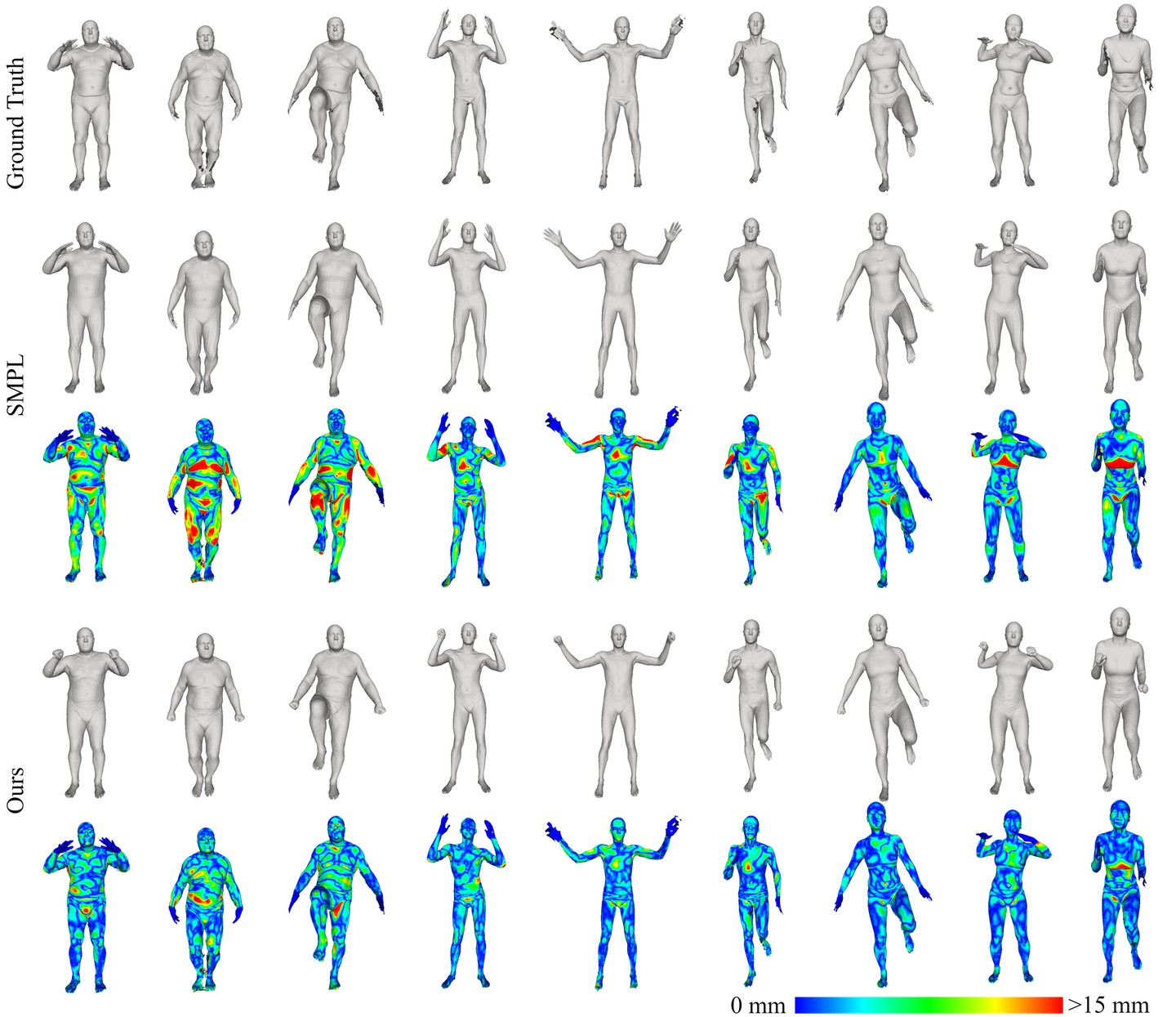}
\end{center}
   \caption{Reconstruction results of our method and SMPL on scan data of DFaust~\cite{bogo2017dynamic}. The error maps show that our method has better reconstruction accuracy.}

\label{fig:DFaust_dense_err}
\end{figure*}

\textbf{Reconstruction from Our Test Dataset.} We compare the reconstruction capability of our method, Baseline, and meshVAE on our test dataset. We obtain the embedding by solving Eq.~\eqref{equ:decoder} for each method.  Tab.~\ref{tab:test_recons} reports the MED errors, and Fig.~\ref{fig:test_rec} visualizes the reconstruction results and their respective error maps.
It can be seen that our method outperforms Baseline and meshVAE. In particular, the MED of our method is lower than that of other methods, which demonstrates the effectiveness of our disentangled and hierarchical architecture design.

\textbf{Reconstruction from Shape Scan Data.} We now test the performance of reconstruction from scan data of human bodies with different identities. We use highly accurate scan data of six males and females with varying body shapes at the same neutral pose. These subjects are irrelevant to our train dataset, and all wear tight clothes. The scan system includes 4 Xtion sensors. When we scan a subject, the subject stands in the center of the scene, and four sensors rotate around the subject. We use the collected multiview RGB and depth data to recover the high accuracy geometry of the subject.

We label eight corresponding landmarks on the scans and use this sparse correspondence to generate coarse alignment with scan data. Then we use point-to-plane iterative closest point (ICP) optimization. For our method, Baseline, and meshVAE, we use the latent parameter regularization of Eq.~\eqref{equ:decoder}. As for the optimization of SMPL, SMPL-X, and Adam models, we adopt the pose prior from~\cite{bogo2016keep} and the shape regularization to constrain their parameters. All the optimizations are implemented based on the Adam method with PyTorch.

We compute PMD for each point in the scan data, and draw the Cumulative Errors Distribution (CED) curve in Fig.~\ref{fig:Apose_curve}. Tab.~\ref{tab:Apose_table} gives numerical comparison and Fig.~\ref{fig:Apose_errormap} shows two examples on the shape scan dataset. Again, our method has the best reconstruction accuracy.

\begin{table}
\begin{center}
\caption{Quantitative comparison of different methods on DFaust~\cite{bogo2017dynamic} scan dataset. The mean PMD ($mm$), the standard deviation, and the valid number of points for testing (without hand part) are reported. We also give the errors (wh) for all points in the last two columns just for reference, although our method does not consider the hand part.}
\label{tab:DFaust_dense}
\begin{tabular}{|c|c|c|c|c|c|}
\hline
Methods &mean & std &\#points &mean(wh)&std(wh) \\
\hline
Ours & $\mathbf{2.9}$ & $\mathbf{4.5}$ & 30953504 & $\mathbf{3.6}$ & 7.9 \\
Ours\_s & 3.1  & 4.7 & 30952186 &3.9&8.1 \\
Baseline & 3.3 & 4.8 & 30956373 &4.1&8.2 \\
Baseline\_s & 3.6 & 4.9 & 30956386 &4.4&8.2 \\
SMPL & 4.6 & 5.5 & 31015202 &4.5&$\mathbf{5.8}$ \\
SMPL\_s & 4.8 & 5.8 & 31012640 &4.8&6.1 \\
SMPL-X & 4.8 & 6.8 & 30972558 &4.8&7.2\\
SMPL-X\_s & 4.9 & 6.9 & 30975033&4.9&7.1 \\
meshVAE & 3.2 & 4.6 & 30956136 &4.0& 8.0 \\
Adam & 14.2 & 15.9 & 30956046&14.0&15.9 \\
Adam\_s & 14.2 & 15.9 & 30956413&14.0 &15.9 \\
\hline
\end{tabular}
\end{center}
\vspace{-0.1in}
\end{table}

\begin{figure}
\begin{center}
\includegraphics[width=\linewidth]{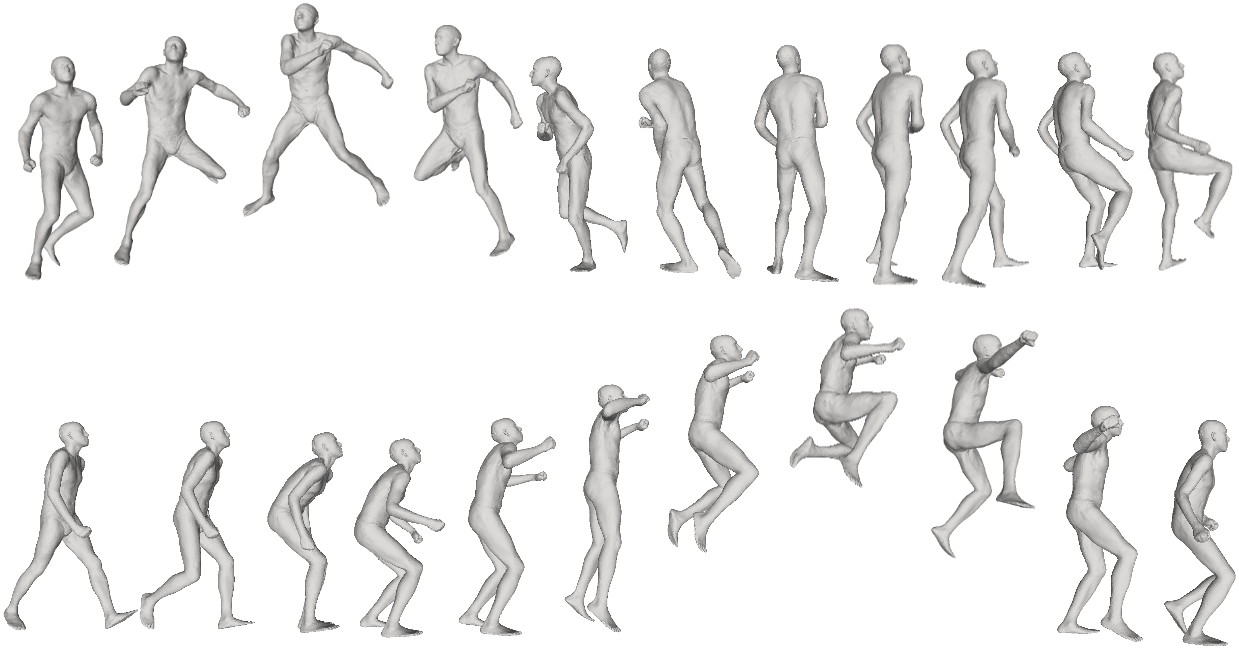}
\end{center}
   \caption{Reconstruction examples of our method with sparse marker constraints from two CMU MOCAP sequences.}
\label{fig:mocap}
\end{figure}

\begin{table}
\begin{center}
\caption{Quantitative comparison of sparse reconstruction on DFaust~\cite{bogo2017dynamic} scan dataset with different methods. The mean PMD($mm$), standard deviation, and the valid number of points for testing (without hand part) are reported.}
\label{tab:DFaust_sparse}
\begin{tabular}{|c|c|c|c|}
\hline
Methods &mean & std & \#points \\
\hline
Ours & 6.3 & 6.4 & 30962584 \\
Ours\_s & $\mathbf{6.2}$  & $\mathbf{6.3}$ & 30963799 \\
SMPL & 6.9 & 7.5 & 31017249 \\
SMPL\_s & 6.7 & 7.2 & 31018236 \\
\hline
\end{tabular}
\end{center}
\vspace{-0.1in}
\end{table}

\textbf{Reconstruction from General Scan Data.} We run the reconstruction for different poses using our method and the baselines on DFaust~\cite{bogo2017dynamic} dataset. DFaust provides ten subjects with several sequences of motion scan, represented as registered meshes. DFaust contains a few subjects that are also in our train dataset Dyna, Faust, or MANO. We choose three subjects from DFaust, labeled with 50007, 50009, and 50020, as our test set. We remove those subjects, which appear in DFaust,  from our train set, and use 1973 poses and 3021 neutral shapes to re-train our model, Baseline model, and meshVAE. We sample data from DFaust with 40 frames interval and finally obtain 108, 65, and 69 test data sets for the three subjects, respectively.

We use the similar point-to-plane ICP registration method with 79 sparse landmarks to carry out a coarse alignment for general pose scan data. For methods that disentangle shape and pose, we perform another optimization by sharing shape parameters among all scan data of one subject, and we denote this approach by a suffix $s$ in the method's name.

We compute PMD for each point in the scan data and draw the Cumulative Errors Distribution (CED) curve in Fig.~\ref{fig:DFaust_dense}. Tab.~\ref{tab:DFaust_dense} gives numerical comparison. Fig.~\ref{fig:DFaust_dense_err} presents several sets of scan data, the reconstructed meshes of our method and SMPL, and the error maps on scan point clouds. It can be seen that our method has the best reconstruction accuracy, and Ours\_s achieves the second-best reconstruction accuracy. The results indicate that our method can effectively disentangle shape and pose variations of a human body.

\begin{table}
\begin{center}
\caption{Quantitative evaluations of 3D pose estimation on H3.6M~\cite{ionescu2013human3}. Superscripts 1 and 2 stand for ground truth and estimated 2D joints input, respectively. Ours\_e is our pose expanded model. The error is the mean Euclidean distance(mm) after Procrustes Analysis~\cite{gower1975generalized}.}
\label{tab:H36M}
\begin{tabular}{|c|c|c|c|c|}
\hline
Methods &$Ours^1$ & $Ours\_e^1$ & $Ours\_e^2$ & SMPLify~\cite{bogo2016keep} \\
\hline
Mean & 95.4 & 65.8 & 86.7 &82.3 \\
Median & 89.2  & 55.9 & 76.0 & 69.3\\
\hline
\end{tabular}
\end{center}
\vspace{-0.1in}
\end{table}

\textbf{Reconstruction with Sparse Constraints.} In this experiment, we test our reconstruction with the constraints of sparse marker points. Motion capture systems usually use sparse markers to capture human movements, and thus the ability to reconstruct 3D human body from sparse markers is important. In particular, we perform the test on the selected data of DFaust. We manually mark 39 landmarks in the registered mesh of DFaust, our template, and SMPL template. We use these sparse corresponding landmarks to reconstruct the mesh and compute PMD errors for scan data. Tab.~\ref{tab:DFaust_sparse} shows the numerical results on the test dataset.

 Even without careful optimization for locations and offsets of sparse markers on the human body as Mosh~\cite{loper2014mosh} did, we still get a similar accuracy as SMPL. Moreover, we select two motion sparse marker sequences from CMU MOCAP\footnote{ mocap.cs.cmu.edu} to test our method. Fig.~\ref{fig:mocap} shows the reconstruction results. These experimental results indicate that our latent embedding achieves a reasonable dimensionality reduction for the human body shape manifold and can reproduce plausible human body shape with few markers constraints.

\begin{figure}
\begin{center}
\includegraphics[width=\linewidth]{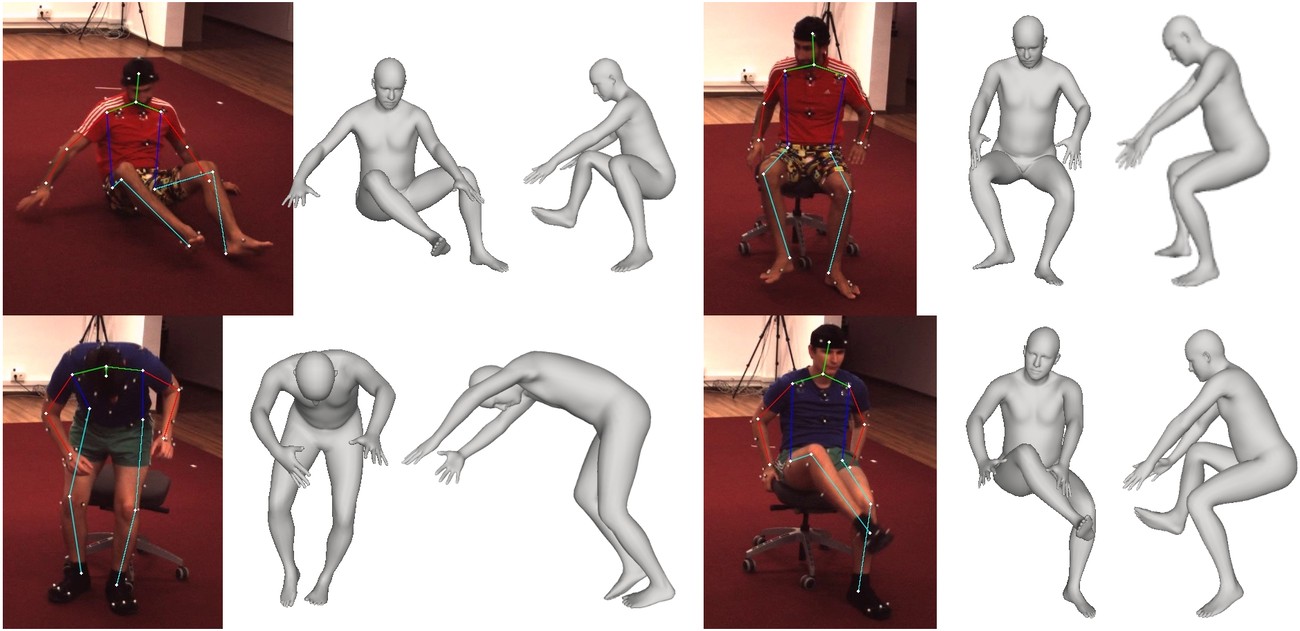}
\end{center}
   \caption{We visualize four worst pose estimation results of $Ours\_e$ model with GT 2D joints input on S9 sitting-down, S9 phoning, S11 smoking, and S11 sitting, respectively. Despite the largest 3D error caused by depth ambiguity, the computed pose still appears plausible.}
\label{fig:h36m}
\end{figure}

\textbf{3D Body Pose Estimation from 2D Joints.}  Although our representation does not define explicit skeleton-like SMPL~\cite{loper2015smpl}, we can also get a rough estimation of a joint by taking the average of manually selected related points on the body mesh. Taking the joint of the elbow as an example, we select vertices around the elbow as the related points. Using this simple strategy, we can generate estimated positions of joints for wrist, knee, and others.

Given 2D human joint positions, we can use our representation to reconstruct the 3D human body model by solving
\begin{equation} \label{equ:2Dpro}
\begin{aligned}
    &\min_{\mathbf{e}_s, \mathbf{e}_p,\mathbf{R},\mathbf{t}}\sum_{joint \text{ } i}\lambda\rho(\Pi_{K}(\mathbf{R} {J}_i(\mathbf{e}_s, \mathbf{e}_p)+\mathbf{t})-\mathbf{j}_i)\\
     &+ \lambda_{g} E_g(\mathcal{R}(\mathcal{C}(\mathbf{e}_s,\mathbf{e}_p))) + \lambda_{\beta}\|\mathbf{e}_s\|_2^2 +\lambda_{\theta}\|\mathbf{e}_p\|_2^2,
\end{aligned}
\end{equation}
where rotation $\mathbf{R}$ and translation $\mathbf{t}$ are the global rigid transformation parameters, ${J}_i(\mathbf{e}_s, \mathbf{e}_p)$ is $i$-th joint position of the decoded mesh from ${\mathbf{e}_s, \mathbf{e}_p}$, $\mathbf{j}_i$ is the $i$-th 2D joint position, $\Pi_{K}$ is the given camera projection matrix with intrinsic parameters $K$, $\rho$ is the robust differentiable Geman-McClure penalty function~\cite{geman1987statistical} and $\mathcal{R}$ is the operation computing relative rotation of two articulated anatomical components from reconstructed coarse feature $\mathcal{C}(\mathbf{e}_s,\mathbf{e}_p)$. We compute the relative rotations of elbows and knees and use a similar penalty function $E_g$ of~\cite{bogo2016keep} to prevent unnatural bending. We use weights $\lambda$, $\lambda_{g}$, $\lambda_{\beta}$ and $\lambda_{\theta}$ to control the importance of each term in the objective function. In our experiments, we set the values of these weights to 55, 400, 5 and 10 as the default configuration.

\begin{figure}
\begin{center}
\includegraphics[width=\linewidth]{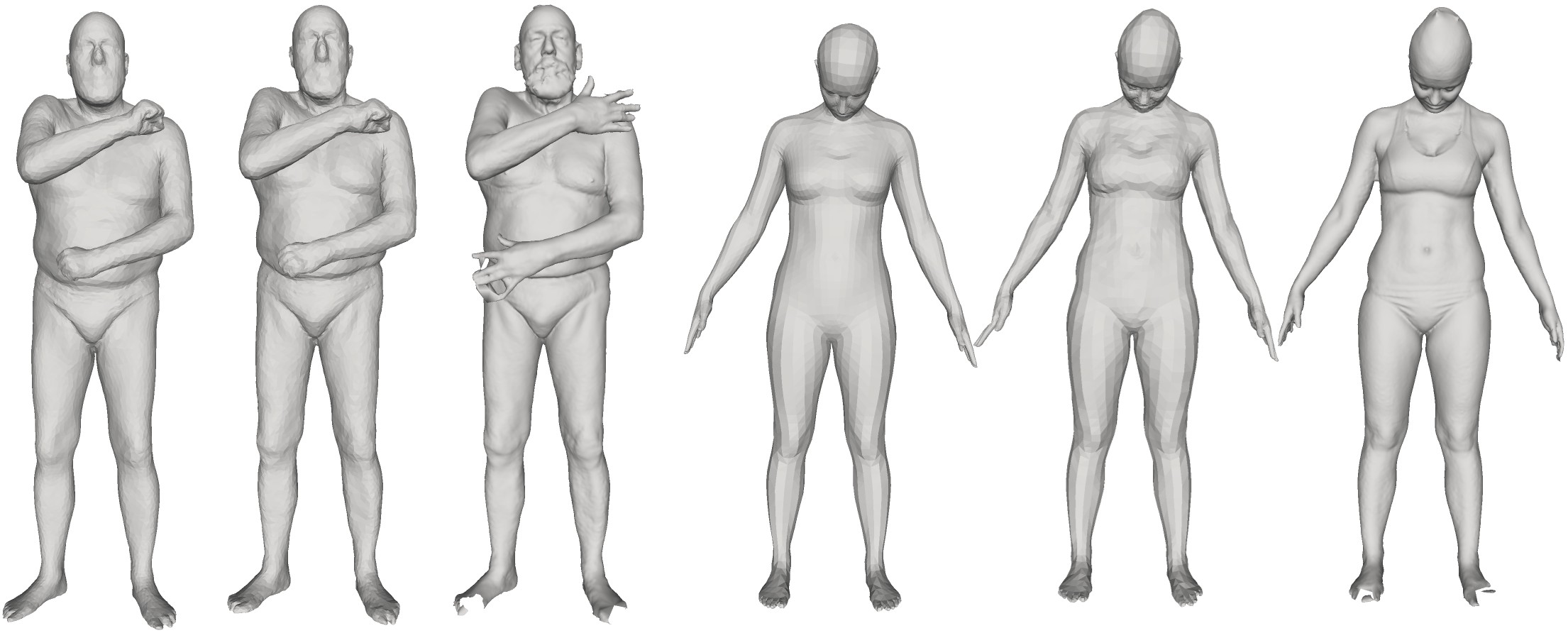}
\end{center}
   \caption{We visualize two registered examples of FAUST with our original model (Ours) and our expanded pose model (Ours\_e), respectively. For each example, from left to right, we show the result meshes expressed by our model, the optimized vertex coordinates, and the scan.}
\label{fig:faust}
\end{figure}

\begin{table}
\begin{center}
\caption{Comparison of the FAUST challenge. AE and WE denote the average and worst errors (cm), respectively. Superscripts 1 and 2 denote the corresponding errors computed with the registered mesh expressed by our model and by the optimized vertex coordinates, respectively. In the table, n.a. indicates that quantitative results are not available. Among the supervised methods, we achieve the best results.}
\label{tab:faust}
\begin{tabular}{|c|c|c|c|c|}
\hline
Methods &Inter AE & Inter WE & Intra AE & Intra We \\
\hline
FMNet~\cite{litany2017deep} & 4.83 & 9.56 & 2.44 &26.16 \\
FARM~\cite{marin2018farm} & 4.12 & 9.98 & 2.81 &19.42 \\
Oshri \etal~\cite{halimi2019unsupervised} & n.a. & n.a. & 2.51 & 24.36  \\
LBS-AE~\cite{li2019lbs} & 4.08 & 10.38 & 2.16 & 6.07  \\
$Ours^1$ & 2.27  & 3.16 & 1.40 & 2.52\\
$Ours^2$ & 2.22  & 3.46 & 1.37 & 3.06\\
$Ours\_e^1$ & 2.52  & 3.59 & 1.17 & 2.39\\
$Ours\_e^2$ & $\mathbf{1.99}$  &$\mathbf{2.99}$ & $\mathbf{1.01}$ & $\mathbf{2.08}$\\
\hline
\end{tabular}
\end{center}
\vspace{-0.1in}
\end{table}

We use an initialization strategy used in SMPLify~\cite{bogo2016keep} and its experiment configuration on H3.6M~\cite{ionescu2013human3}. The only difference is that we compute the results with five frames interval. In Tab.~\ref{tab:H36M}, we give the quantitative results under different configurations.

\begin{figure*}[t]
\begin{center}
\includegraphics[width=\linewidth]{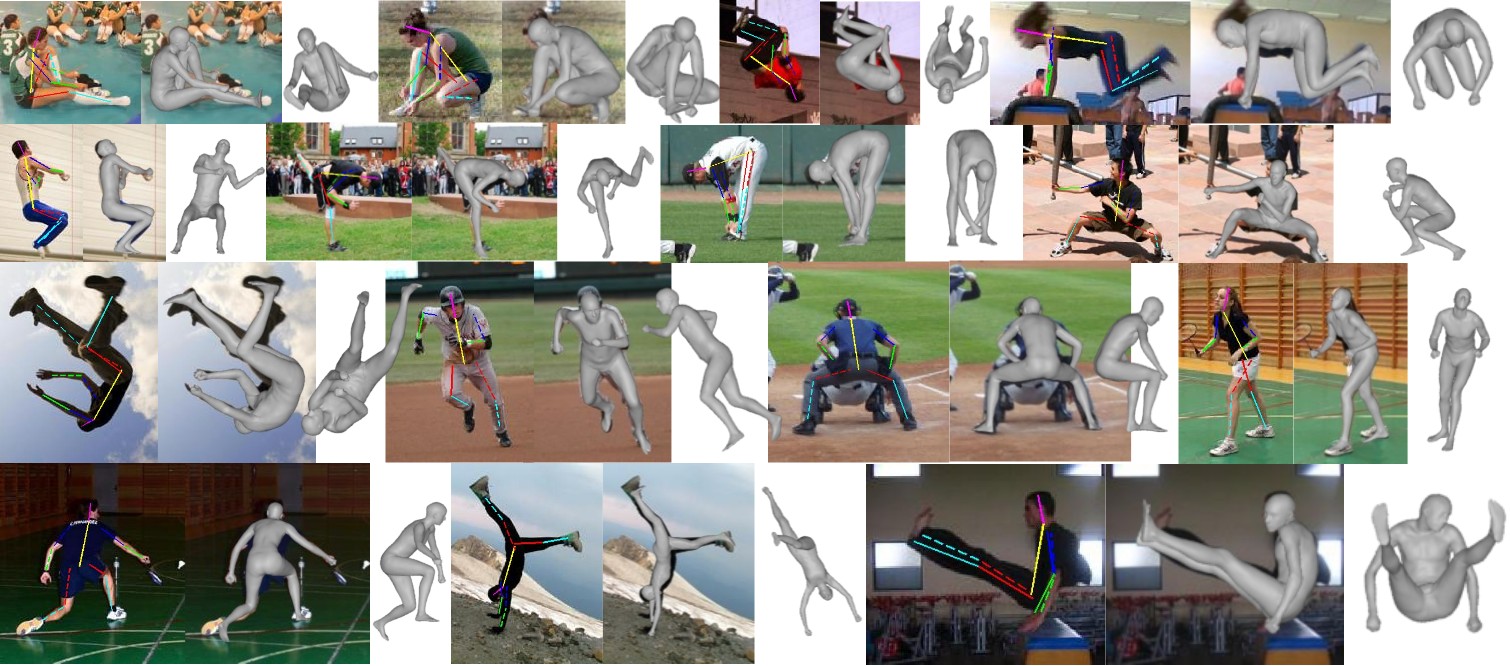}
\end{center}
   \caption{Results of human body estimation on LSP~\cite{johnson2010clustered}. Each group includes the input image with skeleton, the image with estimated human body, and another view of the reconstructed body mesh.}
\label{fig:2Dto3D}
\end{figure*}

First, we use ground truth 2D joints as our input and get a mean error of 95.4mm. We think that more abundant pose data can improve the result because our training pose dataset lacks large-scale actions like sitting down. Therefore, we sample the Moshed CMU dataset~\cite{mahmood2019amass} and use 28600 meshes with abundant poses as our training set to train a new model. As these meshes have the same connectivity with the SMPL template, we use the joints regressor of SMPL to estimate the 3D joints. We use this pose expanded model (Ours\_e) to perform an evaluation on H3.6M with ground truth 2D joints and get a mean error of 65.8mm. In Fig.~\ref{fig:h36m}, we visualize several results of different sequences with the largest error. We can observe that our estimated body pose is reasonable with the 2D joints, even if it has a notable error due to the ambiguity of joint depth.

Then, to compare with SMPLify~\cite{bogo2016keep}, we use its supplied estimated 2D joints as input and get a mean error of 86.7mm. The results of SMPLify is better than ours. However, SMPLify utilizes some prior knowledge like a gender-specific model, specific joints regressor, collision penalty, and a pose prior, while our method does not utilize any prior knowledge except the train data.

We also estimate 3D pose on the LSP~\cite{johnson2010clustered} dataset. Some qualitative results are depicted in Fig.~\ref{fig:2Dto3D}. The results show that our representation can roughly recover the human body from 2D joint locations in images in the wild.

\begin{figure}
\begin{center}
\includegraphics[width=\linewidth]{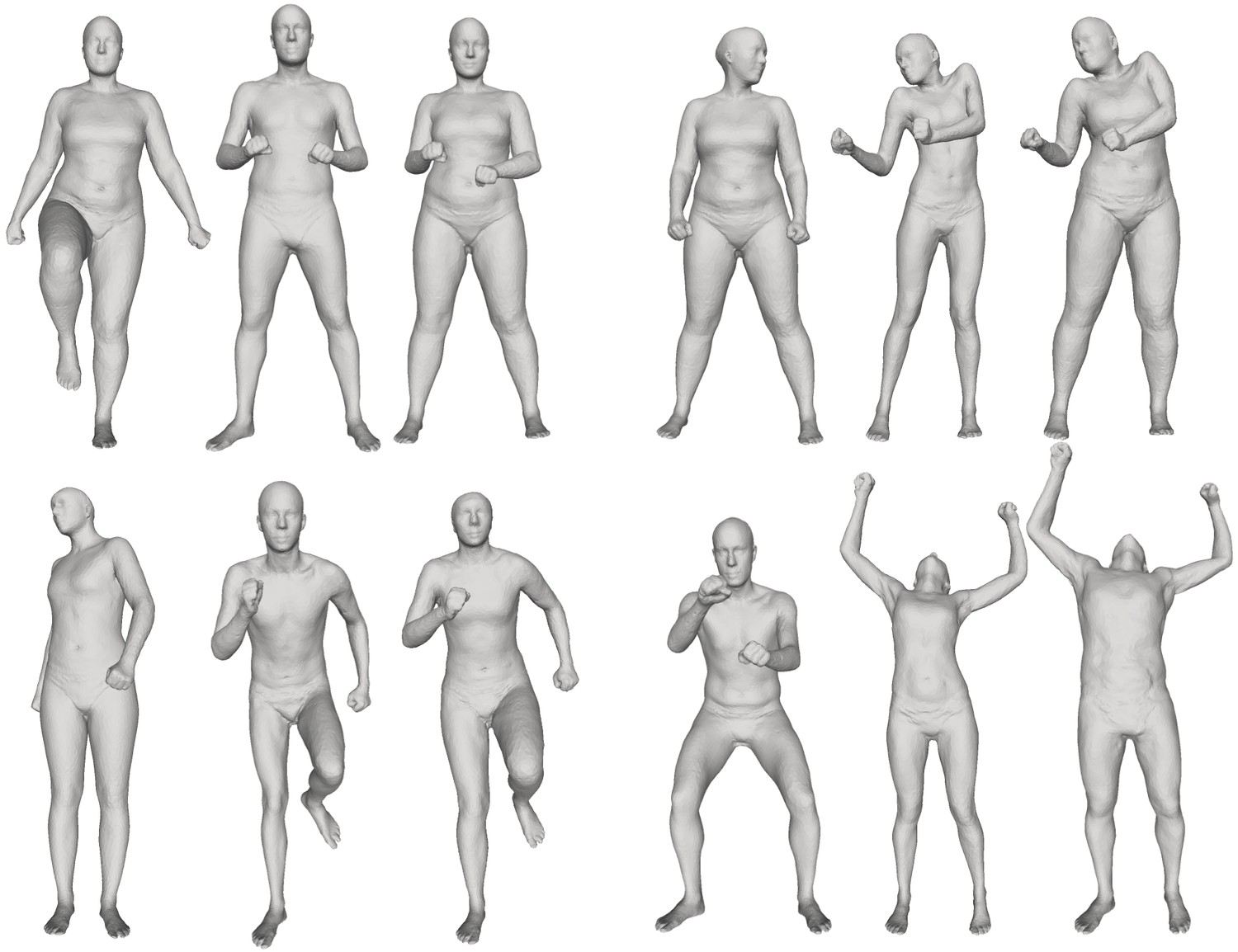}
\end{center}
    \caption{Four pose transfer examples. In each example, there are three meshes. The $\mathbf{e}_s$ of the first mesh and the $\mathbf{e}_p$ of the second mesh  are combined to generate the third mesh (a new body).}
\label{fig:pose_transfer}
\end{figure}

\begin{figure}
\begin{center}
\includegraphics[width=\linewidth]{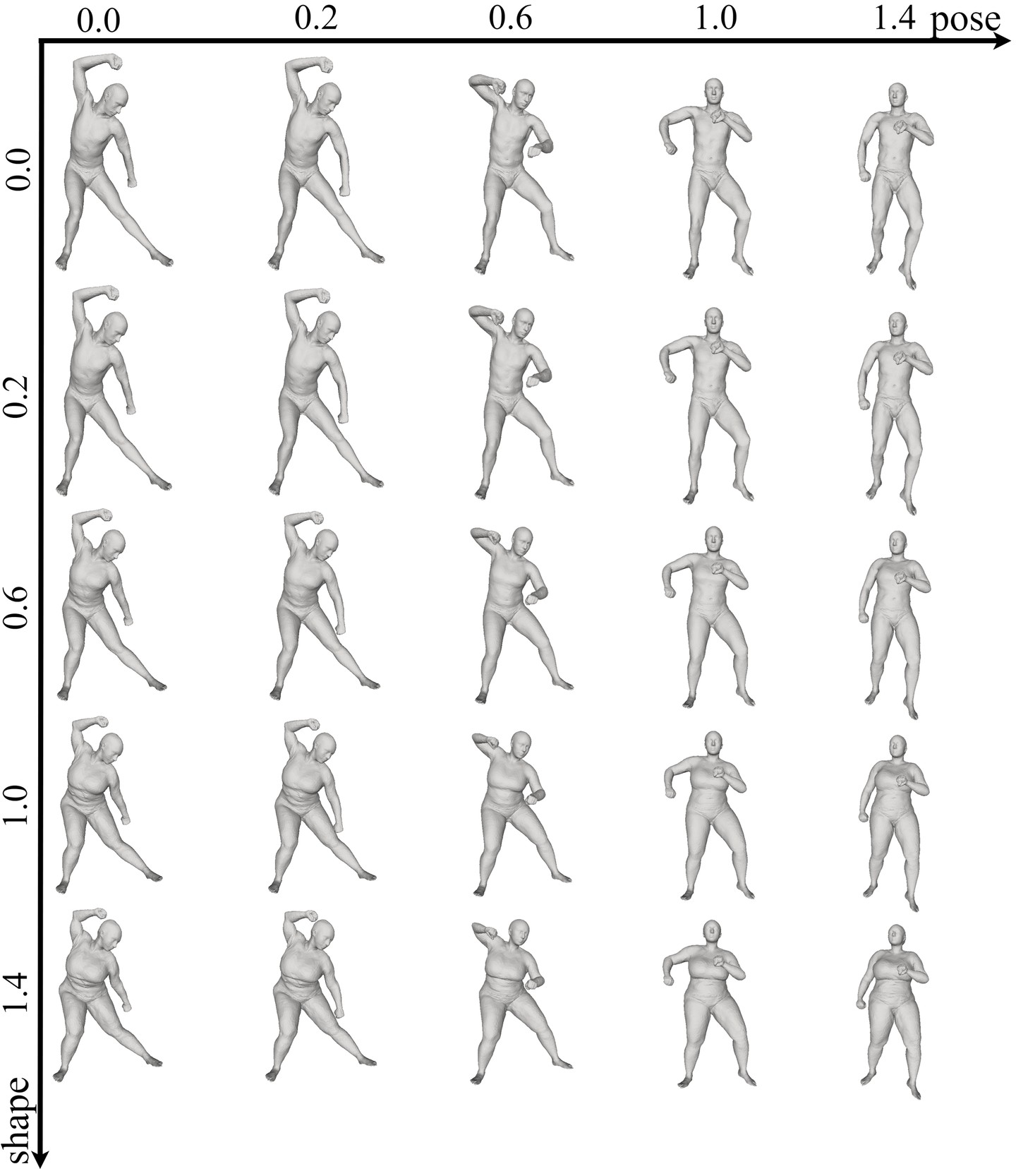}
\end{center}
    \caption{An example of bilinear interpolation.}
\label{fig:inter_meshes}
\end{figure}

\textbf{Performance on FAUST.} In this experiment, we evaluate the alignment of our method on the FAUST benchmark~\cite{bogo2014faust}, which consists of 200 real test scans of human bodies. The ground-truth correspondences of this challenge are not available, and the accuracy evaluation is obtained by submitting correspondence results online.

Given each challenge pair, we use our model to register each scan individually. Then we use the registered meshes as the common domain to establish a point-to-point correspondence between the pair scans.

\begin{figure*}[t]
\begin{center}
\includegraphics[width=\linewidth]{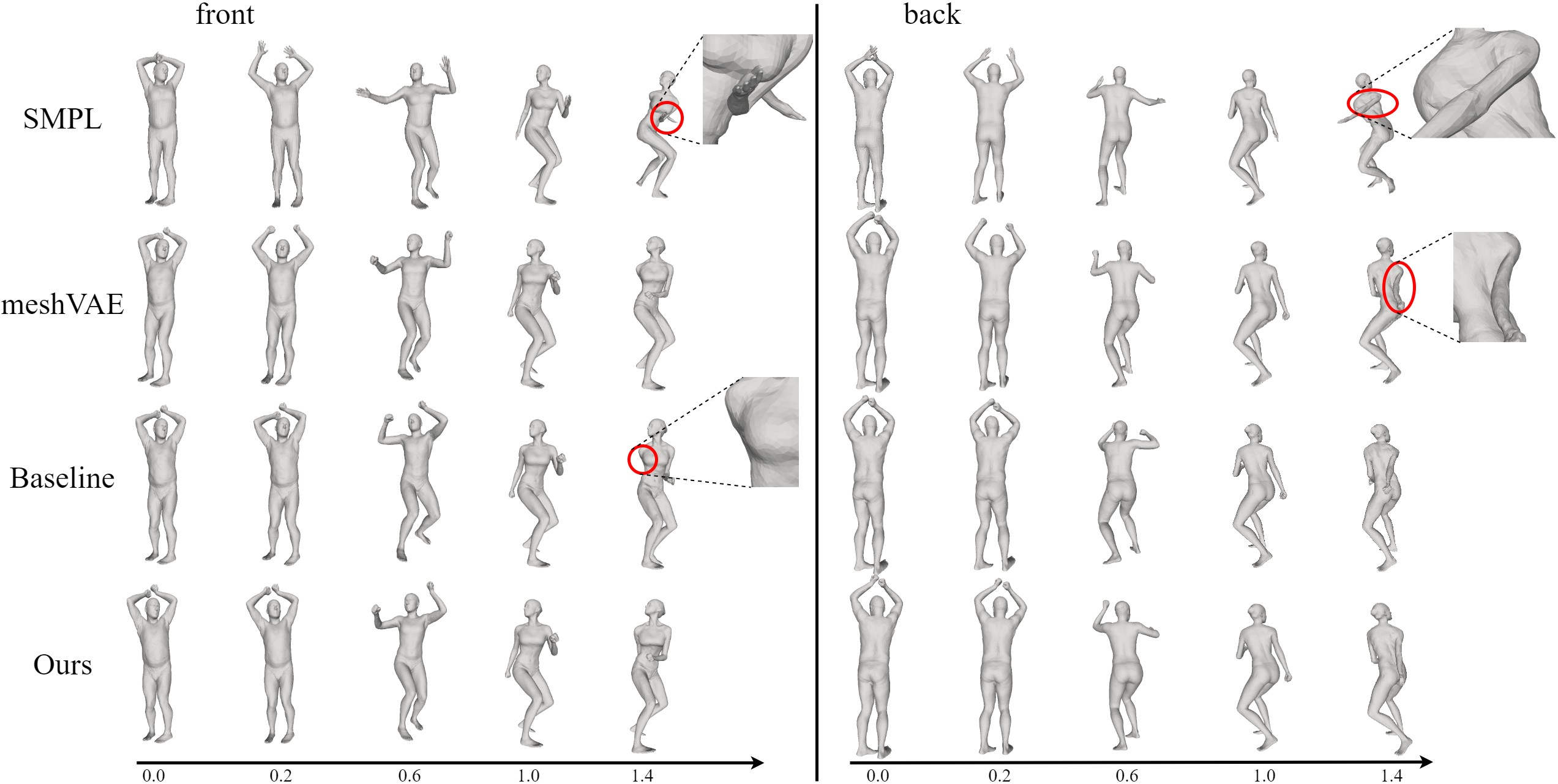}
\end{center}
   \caption{Comparison of four methods for linear interpolation. Red circles indicate unreasonable human body movements.}
\label{fig:compare_vae}
\end{figure*}

We adopt the optimization strategy of the connectivity conversion of Eq.~\eqref{equ:regist}. Instead of constraining the reconstructed mesh within the shape space of our model, we use our model as a geometry prior and introduce free body mesh points as additional optimization parameters. We optimize our model parameters $\{\mathbf{e}_s,\mathbf{e}_p\}$ and the free vertex coordinates simultaneously, and finally get two registered body meshes for each scan. One is expressed by our model directly and the other is the result mesh with optimized vertex coordinates. Fig.~\ref{fig:faust} visualizes two examples of registered results of test scans.

The optimization pipeline described above needs some sparse landmarks as initialization. To test the robustness of our method, we first use only the five landmarks estimated by~\cite{marin2018farm} to do initialization. It works well for poses without large bending of arms and legs, but it cannot generate good registration results for large-scale poses like a deep squat. Therefore, we add another five landmarks in the areas of elbows, knees, and butt. In this way, with totally ten landmarks, we can get correct registration results for all the 200 test scans.

We report the quantitative comparison with the official ranking in Tab.~\ref{tab:faust}. With manually placed landmarks as supervision, our method can achieve the best performance. As our body mesh does not model the hand part, which might result in large errors, we also report the results of Ours\_e model, which achieves better results.

\begin{figure}
\begin{center}
\includegraphics[width=\linewidth]{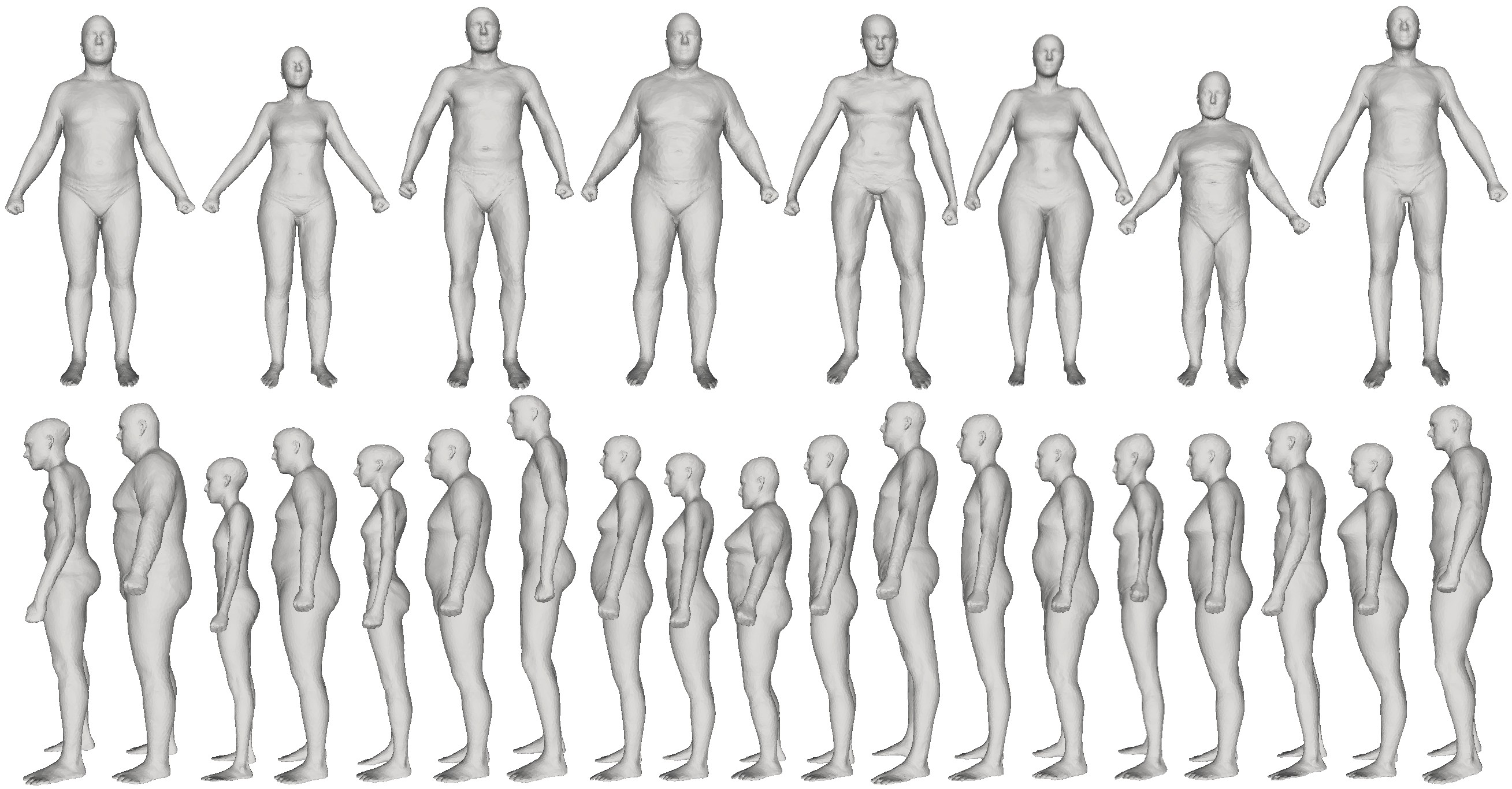}
\end{center}
   \caption{Examples of human bodies with a neutral-pose, generated by randomly sampling shape parameters from $N(0,I)$.} 
\label{fig:sample_shape}
\end{figure}

\begin{figure}
\begin{center}
\includegraphics[width=\linewidth]{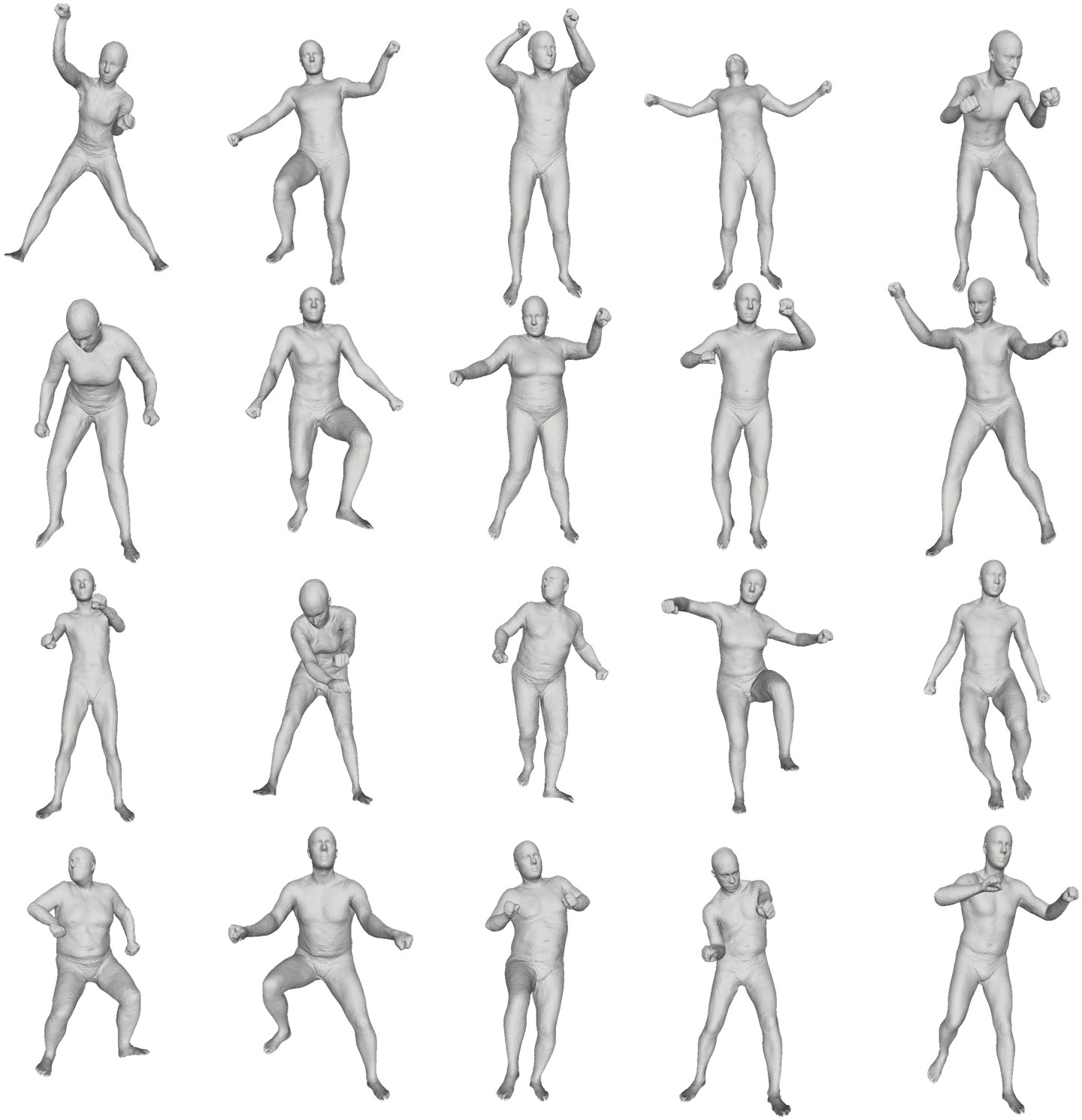}
\end{center}
   \caption{By randomly sampling the shape and pose parameters, we can produce various body shapes with rich posture variations.} 
\label{fig:sample_pose}
\end{figure}

\subsection{Qualitative Evaluation} \label{sec:app}

\textbf{Pose Transfer.} To demonstrate the robustness and disentanglement of our proposed model, we use it for pose transfer by retrieving $\mathbf{e}_s$ of a body mesh and combining it with $\mathbf{e}_p$ from another body mesh to generate a new one. Fig.~\ref{fig:pose_transfer} gives four examples of pose transfer. It can be observed that the generated meshes look natural and have similar pose and identity as the reference meshes.

\begin{figure*}[t]
\begin{center}
\includegraphics[width=\linewidth]{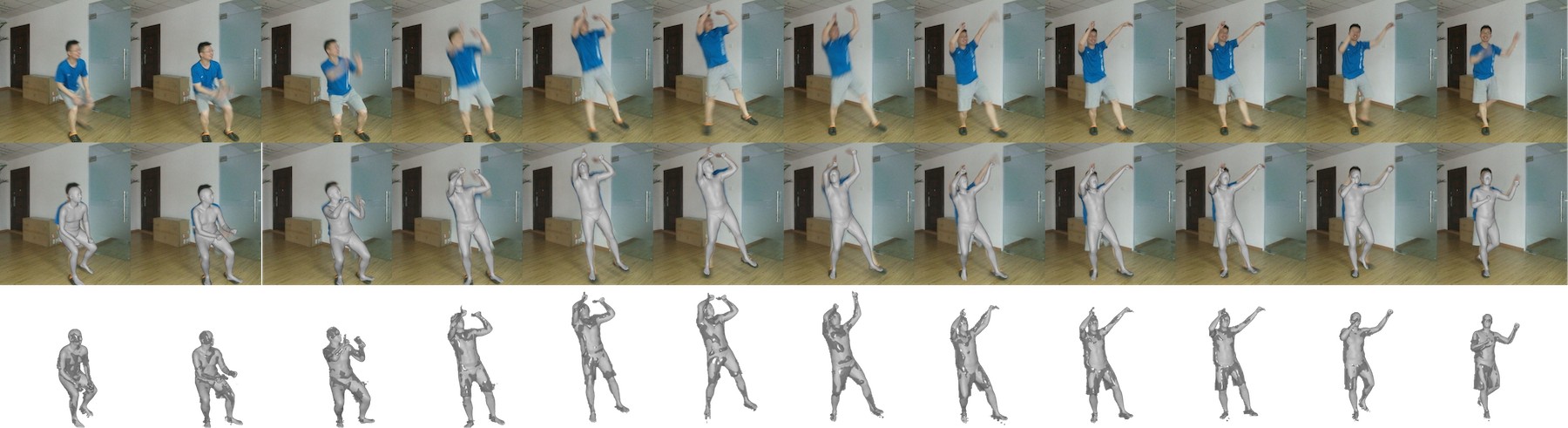}
\end{center}
   \caption{Reconstruction results by fitting our body shape representation to a sequence of depth images. The first row shows the original color images that are not used in our algorithm but just for visualization purpose. The second row shows the registered meshes overlaid on the images. The third row shows the reconstructed meshes and the target point clouds together.}
\label{fig:reg2depth}
\end{figure*}

\textbf{Global Interpolation.} To test the capability of our representation for interpolation between two random people with different poses, we qualitatively compare our method with Baseline, meshVAE, and SMPL. Given the source and target meshes, we first use the reconstruction methods described in Section~\ref{sec:rec_accuracy} to extract the respective parameter values. Then we linearly interpolate between the source and target parameters to generate a list of new parameter values, and finally we use the decoder to construct the meshes.  Fig.~\ref{fig:compare_vae} shows the front view and the back view of the resulting meshes. The four methods produce plausible results for interpolation (i.e., the interpolation parameter lies within [0,1]), but for extrapolation, SMPL generates weird body movements compared to our method. The pose parameters of SMPL record the relative rotation between two joints, which does not consider human body movement prior. This may explain the weird extrapolation results produced by SMPL.

\textbf{Bilinear Interpolation.} Our representation separates shape and pose parameters, which allows us to perform interpolation on each category of the parameters. For example,  given two meshes with different shapes and poses, we first extract their shape and pose parameters, and then we linearly interpolate the shape or pose parameters. Fig.~\ref{fig:inter_meshes} shows the results of such interpolation. We can see that each column has a consistent pose, and each row corresponds to a specific person. Even for extrapolation, the results are reasonably good.

\textbf{New Model Generation.} Since we encode our shape parameters and pose parameters with VAE architecture separately, we can generate new body models by randomly sampling the two sets of embedding parameters.

In Fig.~\ref{fig:sample_shape}, we generate neutral meshes by randomly sampling on the embedded shape space. The generated shapes have abundant variations. In Fig.~\ref{fig:sample_pose}, we randomly generate pose meshes by sampling on the embedded shape and pose spaces. The generated meshes have plausible and different postures.

\textbf{Registration to Depth Images.} We also fit our representation model to a sequence of depth images. Eq.~\eqref{equ:2Dpro} is adapted for this purpose. To smooth the results in the temporal domain, we apply the smooth energy for pose parameters and share one shape parameter for the entire sequence. We use Kinect v2 to collect depth data. For each frame, we convert the depth image to a mesh for the convenience of point-to-plane ICP registration. Besides the depth data, we also use the 3D joint locations predicted by the SDK of Kinect v2. The prediction is not very accurate. It just provides a coarse initialization.  Fig.~\ref{fig:reg2depth} shows an example of such registration to a sequence of depth images, where the color images are not used in our algorithm and just for visualization.

\section{Limitations}
Our work has several limitations.
First, while our representation defines a coarse-level shape, it lacks an explicit and simple method for computing the position of body skeleton from latent embedding. To estimate a joint of the skeleton, currently we just average the positions of those mesh vertices related to the joint. This estimation, however, is not very accurate and may introduce errors into the target human pose.

Second, for the neutral pose, we directly use the common pose of SPRING~\cite{yang2014semantic}. Nevertheless, the postures of SPRING are not totally consistent. There exist small misalignments in this dataset. For example, arms may have small swings, and heads may have some deviations in their orientation. These misalignments affect learning accuracy.

Third, as the ACAP feature used in the algorithm is defined on triangular meshes, our method cannot be directly applied to other data types such as voxels or point clouds.

Fourth, our method requires mesh data to have consistent connectivity. To use or handle scan data, usually we need to perform time-consuming connectivity conversion. A possible solution to avoid the connectivity conversion is to adopt a self-supervised training loss and a discriminator on decoded ACAP feature. We will explore this problem as future work.

\section{Conclusions}
We have presented a general framework for learning and reconstructing 3D human body models. A VAE like architecture is used to learn disentangled human body shape and pose embedding and train our model end-to-end. A coarse-to-fine pipeline is proposed to reconstruct high accurate body models.
To make full use of the great fitting ability of neural network, we have constructed a large dataset consisting of models with consistent connectivity. These models are represented by neutral shapes corresponding to their identities and deformation information for individual shape variations. Experimental results have demonstrated the advantages of our learned embedding in terms of the accuracy of reconstruction and the flexibility for model recreation. The trained model and the constructed dataset will be made publicly available. We believe that the learned embedding and dataset will be useful for various human body related applications.

\section*{Acknowledgments}
We thank VRC Inc. (Japan) for sharing the scanned human shape models with us in Fig.~\ref{fig:Apose_errormap} and Tab.~\ref{tab:Apose_table}. This research is partially supported by National Natural Science Foundation of China (No. 61672481), Youth Innovation Promotion Association CAS (No. 2018495), NTU Data Science and Artificial Intelligence Research Center (DSAIR) (No. M4082285), MoE Tier-2 Grant (2016-T2-2-065, 2017-T2-1-076) of Singapore, and the National Research Foundation, Singapore under its International Research Centres in Singapore Funding Initiative. Any opinions, findings, and conclusions or recommendations expressed in this material are those of the authors and do not reflect the views of the National Research Foundation, Singapore.

{\small
\bibliographystyle{ieee}
\bibliography{egbib}
}

\section*{Appendix}
\subsection*{Network Details.} In our network, all basic transformation modules except for the learnable skinning layer are designed as an MLP, which is a stack of a unit structure. The unit structure is composed of a fully connected layer, followed by a $tanh$ activation function. Tab.~\ref{tab:architecture} gives the detailed information of each MLP in the decoder.

\begin{table}[h!]
\begin{center}
\caption{Structure information of MLPs in the decoder, which includes the number of the composed units of each MLP and the dimensions of the input and output feature of each stacked unit in the MLP.}
\vskip -0.5cm
\label{tab:architecture}
\begin{tabular}{|c|c|c|c|c|c|}
\hline
MLP & $\mathcal{C}_s$ \& $\mathcal{D}_s$ & $\mathcal{C}_p$ \& $\mathcal{D}_p$  & $\mathcal{T}_c$ & $\mathcal{T}_d$\\
\hline
units number& 2 &2  &1 & 1 \\
dimensions & $50,400,800$ &$72,400,800$  &$800,144$ & $800,9|\mathcal{V}|$ \\
\hline
\end{tabular}
\end{center}
\vspace{-0.1in}
\end{table}

\subsection*{Training Details.} The hyperparameters $\lambda_s$, $\lambda_p$, $\lambda_{r_1}$, $\lambda_{r_2}$, $\lambda_{r_{c_1}}$ and $\lambda_{r_{c_2}}$ in Eq.~\eqref{eq:loss} control the trade-off between the KL loss and the reconstruction loss. To find the optimal configuration, we fix $\lambda_s$ and $\lambda_p$ to 1 and adjust the reconstruction related hyperparameters. In Tab.~\ref{tab:ablation}, we show the ablation study of these parameters. To balance the KL loss and the reconstruction loss, we use the second configuration to train for about 1600 epochs, and then fine-tune the trained model with the first configuration for another 600 epochs. We set the batch size to 24. Each batch is composed of two sets of data of equal amounts from the Neutral and Pose datasets. The learning rate is set to $1.0\times10^{-4}$. The entire training can be completed in less than 15 hours on a single NVIDIA TITAN Xp GPU.

Unlike \cite{tan2017variational,tan2017mesh}, we use $\ell_{1}$ instead of $\ell_{2}$ as the reconstruction loss because we find that the $\ell_{2}$ loss often results in a higher KL loss to achieve equivalent feature reconstruction accuracy. The last row of Tab.~\ref{tab:ablation} shows the optimal test error of the model trained with the $\ell_{2}$ loss.

\begin{table}[h!]
\begin{center}
\caption{The ablation study of hyperparameters in training loss. We report the loss on the test dataset of models trained with different parameter settings. For the $\ell_{1}$ training loss, we fix the KL loss weights $\lambda_s$ and $\lambda_p$ to 1.0, and set the ratios of $(\lambda_{r_1}:\lambda_{r_{c_1}})$ and $(\lambda_{r_2}:\lambda_{r_{c_2}})$ to (1:0.6). We test different values for the two groups of parameters. In the last row, we report the optimal test error of the model trained with the $\ell_{2}$ loss. Its KL loss is larger than the results trained by the $\ell_{1}$ loss when they achieve equivalent test accuracy.}
\vskip -0.5cm
\label{tab:ablation}
\begin{tabular}{|c|c|c|c|c|c|c|}
\hline
$\lambda_{r_1},\lambda_{r_2}$ & $E_{sKL}$ & $E_{pKL}$  & $E_{L1_1}$ & $E_{L1_2}$ &$E_{L1_{c_1}}$ & $E_{L1_{c_2}}$\\
\hline
1e3,1e4 & 73.51 & 17.30 & 0.063 & 0.055 & 0.062 & 0.049 \\
2.5e3,2.5e4 & 94.97 & 86.97 & 0.035 & 0.053 & 0.031 & 0.047 \\
5e3,5e4 & 150.30 & 147.36 & 0.029 & 0.054 & 0.026 & 0.047 \\
1e4,5e4 & 199.96 & 280.00 & 0.021 & 0.054 & 0.019 & 0.047 \\
\hline
 $\ell_{2}$ & 126.85 & 168.38 & 0.031 & 0.055 & 0.028 & 0.048 \\
\hline
\end{tabular}
\end{center}
\vspace{-0.1in}
\end{table} 

\end{document}